\pdfoutput=1

\documentclass[11pt]{article}

\usepackage[final]{acl}

\usepackage{times}
\usepackage{latexsym}
\usepackage[breakable]{tcolorbox}

\usepackage[T1]{fontenc}

\usepackage[utf8]{inputenc}

\usepackage{microtype}

\usepackage{inconsolata}
\usepackage{booktabs}

\usepackage{fontawesome}
\usepackage{graphicx}
\usepackage{amsmath}
\usepackage{amssymb}
\usepackage{booktabs}

\newcommand{\dataset}[1]{\textsf{DATASET}}

\title{Weaving Context Across Images: Improving Vision-Language Models through Focus-Centric Visual Chains}

\author{Juntian Zhang \\
  Affiliation / Address line 1 \\
  Affiliation / Address line 2 \\
  Affiliation / Address line 3 \\
  \texttt{email@domain} \\\And
  Chuanqi cheng \\
  Affiliation / Address line 1 \\
  Affiliation / Address line 2 \\
  Affiliation / Address line 3 \\
  \texttt{email@domain} \\}

\author{
   Juntian Zhang\textsuperscript{1}, Chuanqi Cheng\textsuperscript{1}, Yuhan Liu\textsuperscript{1}\thanks{\ \ Corresponding authors.}, Wei Liu\textsuperscript{2}, Jian Luan\textsuperscript{2},
  Rui Yan\textsuperscript{1,3}\footnotemark[1] \\
  \textsuperscript{1}Gaoling School of Artificial Intelligence, Renmin University of China, \textsuperscript{2}Xiaomi, \textsuperscript{3}Wuhan University\\
  \texttt{zhangjuntian@ruc.edu.cn}
}

\begin{document}
\maketitle
\begin{abstract}

Vision-language models (VLMs) achieve remarkable success in single-image tasks. However, real-world scenarios often involve intricate multi-image inputs, leading to a notable performance decline as models struggle to disentangle critical information scattered across complex visual features.
In this work, we propose \textit{\textbf{Focus-Centric Visual Chain}}, a novel paradigm that enhances VLMs’ perception, comprehension, and reasoning abilities in multi-image scenarios. To facilitate this paradigm, we propose \textit{\textbf{Focus-Centric Data Synthesis}}, a scalable bottom-up approach for synthesizing high-quality data with elaborate reasoning paths. Through this approach, We
construct \textbf{VISC-150K}, a large-scale dataset with reasoning data in the form of \textit{Focus-Centric Visual Chain}, specifically designed for multi-image tasks. Experimental results on seven multi-image benchmarks demonstrate that our method achieves average performance gains of \textbf{3.16\%} and \textbf{2.24\%} across two distinct model architectures, without compromising the general vision-language capabilities.
Our study represents a significant step toward more robust and capable vision-language systems that can handle complex visual scenarios:
\faGithub~\href{https://anonymous.4open.science/r/VISC-2013}
{VISC}.

\end{abstract}

\section{Introduction}

The rapid advancement of VLMs has revolutionized traditional visual tasks with single-image input, achieving human-level performance in various applications~\cite{daniali2023perception}. However, real-world scenarios frequently involve more complex visual input, such as multiple images, where current VLMs show significant performance degradation~\cite{zhao2024benchmarking}.
The challenges stem from two complementary traits of multi-image tasks:
(1) \textbf{Cross-image correlations}: Images are often diversely related, requiring a holistic understanding of their contextual relationships.
(2) \textbf{Visual discontinuity}: The fragmentation of information between images makes it challenging to accurately grasp cross-image relationships.

\begin{figure}[t]
    \centering
    \includegraphics[width=\columnwidth]{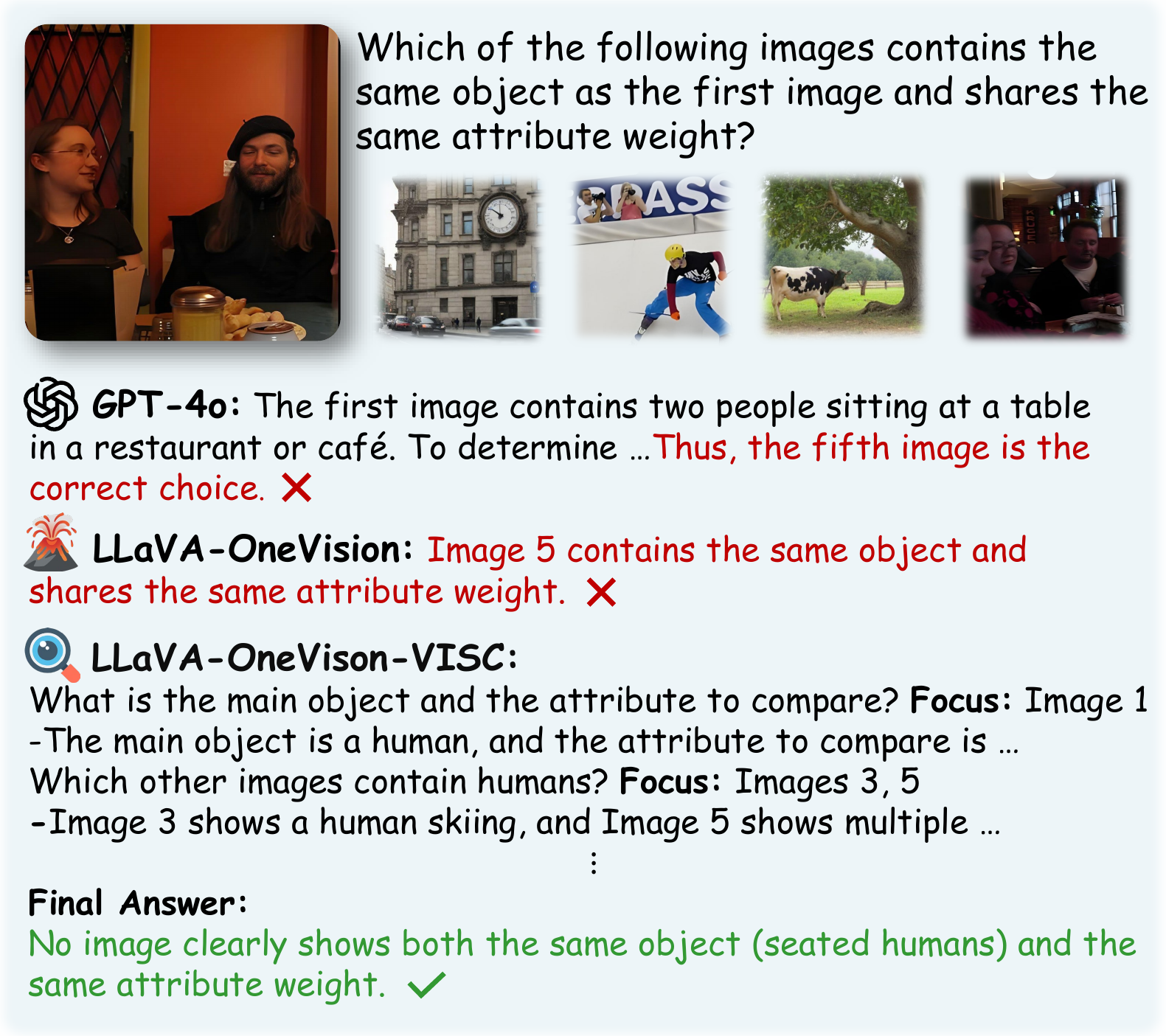}
    \caption{
    A multi-image QA example: Using \textit{Focus-Centric Visual Chain}, our model LLaVA-OneVision-VISC successfully answers a question that both GPT-4o and LLaVA-OneVision fail to solve correctly. 
    }
    \label{case}
\end{figure}

To address these challenges, we propose \textbf{\textit{Focus-Centric Visual Chain}}, a multi-image reasoning paradigm that progressively focuses on key information. In this process, VLMs decompose a complex task into a sequence of sub-processes, each involving the resolution of a sub-question that focuses on localized visual input. As illustrated in Figure~\ref{case}, this process allows models to iteratively aggregate the necessary visual evidence guided by the sub-questions, ultimately leading to the resolution of the complex task.

To implement this paradigm, high-quality reasoning data is indispensable. However, despite extensive research on reasoning tasks, reasoning data for multi-image scenarios remains scarce. While there exist approaches leveraging multimodal models to directly generate reasoning chains~\cite{zhang2023multimodal}or distilling data from more muscular models, such methods present two primary limitations: (1) \textbf{Insufficient reliability}, even state-of-the-art closed-source models (e.g., GPT-4o) demonstrate inconsistent performance on multi-image tasks~\cite{wang2024muirbench}; and (2) \textbf{Prohibitive costs}, the substantial cost of closed-source models severely constrains scalability. Hence, we propose the \textbf{\textit{Focus-Centric Data Synthesis}} framework, an efficient approach for generating reasoning data following a bottom-up strategy. In contrast to our reasoning paradigm, which decomposes complex tasks into simpler sub-tasks, the framework centers on progressive information aggregation. At each stage, it expands the existing information set to formulate reliable reasoning paths and corresponding questions from complex visual inputs.

The \textit{Focus-Centric Data Synthesis} framework comprises four systematically designed modules: (1) \textit{Feature Extraction} constructs comprehensive textual profiles for each image, which serve as nodes in subsequent process; (2) \textit{Pair Connection} identifies relevant image pairs through object-oriented and event-oriented detection, forming edges between distinct nodes; (3) \textit{Relevance Annotation} categorizes detected connections into three predefined types (Temporal, Spatial, and Semantic) and detail them; (4) \textit{Question Generation} produces logically chained sub-questions based on the established inter-image network, ultimately synthesizing the final composite question and reasoning path. 
The framework's bottom-up design ensures data quality while maintaining computational efficiency through exclusive use of open-source models.

Leveraging this framework, we construct \textbf{VISC-150K} composed of 150K high-quality multi-image reasoning samples. Extensive experiments across seven multi-image benchmarks demonstrate the effectiveness of our approach. When integrated with different base models, VISC-150K consistently brings performance improvements across all challenging benchmarks, with average accuracy increased by 3.16\% and 2.24\%, respectively, achieving new state-of-the-art on four out of the seven.

In summary, our contributions are three-fold:
\begin{itemize}
\item We introduce the \textit{Focus-Centric Visual Chain} paradigm to solve complex multi-image tasks through question decomposition and stepwise reasoning.
\item We propose \textit{Focus-Centric Data Synthesis}, a framework tackling data scarcity by synthesizing reliable, cost-effective, and reproducible reasoning data via open-source models.
\item We release VISC-150K, a multi-image reasoning dataset containing 150K data with Focus-Centric Visual Chains. Our dataset delivers consistent performance gains across diverse model architectures on seven challenging multi-image benchmarks, as validated by comprehensive evaluations.
\end{itemize}

\section{Related Work}

\subsection{Vision-Language Models}
Vision-Language Models (VLMs) integrate visual and textual processing through multimodal architectures, empowering various types of tasks~\cite{liu2024skepticism, liu2024tiny}. Closed-source models such as GPT-4o~\cite{openai2024gpt4o} and Gemini-1.5-Pro~\cite{gemini2024} demonstrate state-of-the-art performance. While open-source VLMs can be categorized into two types based on their architecture.
The first type employs a unified architecture for both visual and textual modality. For example, Flamingo~\cite{alayrac2022flamingovisuallanguagemodel} incorporates visual information into textual inputs through blocks based on cross-attention. The KOSMOS series~\cite{huang2023languageneedaligningperception,peng2023kosmos2groundingmultimodallarge} uses the same embedding module to encode text and visual information. 

The second type aligns the two modalities by utilizing a connector module to project visual inputs into the textual space. 
BLIP-2~\cite{li2023blip2} connects the visual encoder and the language model with Q-former. 
InstructBLIP~\cite{dai2023instructblip} further proposes an innovative instruction integration to achieve better modality alignment.
The LLaVA series~\cite{liu2023visual} adopts a more concise design, using MLP as the mapping layer between modalities. 
This architecture was widely adopted by recent VLMs, including Mantis~\cite{jiang2024mantis}, LLaVA-OneVision~\cite{li2024llavaonevision}, and InternVL2~\cite{chen2024far}. Additionally, Qwen2-VL~\cite{Qwen2VL} adopts visual position encoding and Naive Dynamic Resolution.

Despite advancements, evaluations reveal two critical limitations of previous works: (1) VLMs struggle with complex multi-image tasks~\cite{zhao2024benchmarking}
and (2) existing training paradigms offer diminishing returns for multi-image tasks~\cite{campbell2024understanding}. To overcome these limitations, we synthesize a dataset that continuously improves VLMs' performance on various multi-image benchmarks, thus providing a new solution for complex multi-image scenarios.

\subsection{Reasoning of LLMs and VLMs}

Reasoning capabilities in Large Language Models (LLMs) have evolved through innovative prompting strategies and knowledge integration. Chain-of-Thought (CoT)~\cite{wei2022chain} pioneered reasoning in LLM, later enhanced by Tree-of-Thought\cite{yao2023tree} and Graph-of-Thought~\cite{besta2023graph}. Self-consistency~\cite{wang2023self} improves robustness by aggregating multiple reasoning paths via voting strategy. 
Recent advances like OpenAI’s o1 series~\cite{openai2024o1} employ reinforcement learning to foster more complex reasoning.

Beyond focusing on single-modal, multimodal reasoning~\cite{wang2024exploringreasoningabilitiesmultimodal} extensions leverage data-centric approaches~\cite{gao2023gllava,zhang2024mavis,shi2024mathllava,cheng2024least,xu2024llava}, knowledge graph integration~\cite{zhang2024multi,lee2024multimodal}, and tree search~\cite{yao2024mulberry} for transferring reasoning capabilities to VLMs. 
However, for scenarios with multi-image inputs, how to enable VLMs to integrate complex visual information and perform slow thinking remains an issue that has not been fully explored. To address this issue, we design a multi-step reasoning paradigm, decomposing the initial question into a sequence of sub-questions and focusing on a subset of input images in each step. The effectiveness of this paradigm is verified by experiments in \S~\ref{exp}.
\begin{figure*}[t]
    \centering
    \includegraphics[width=\textwidth]{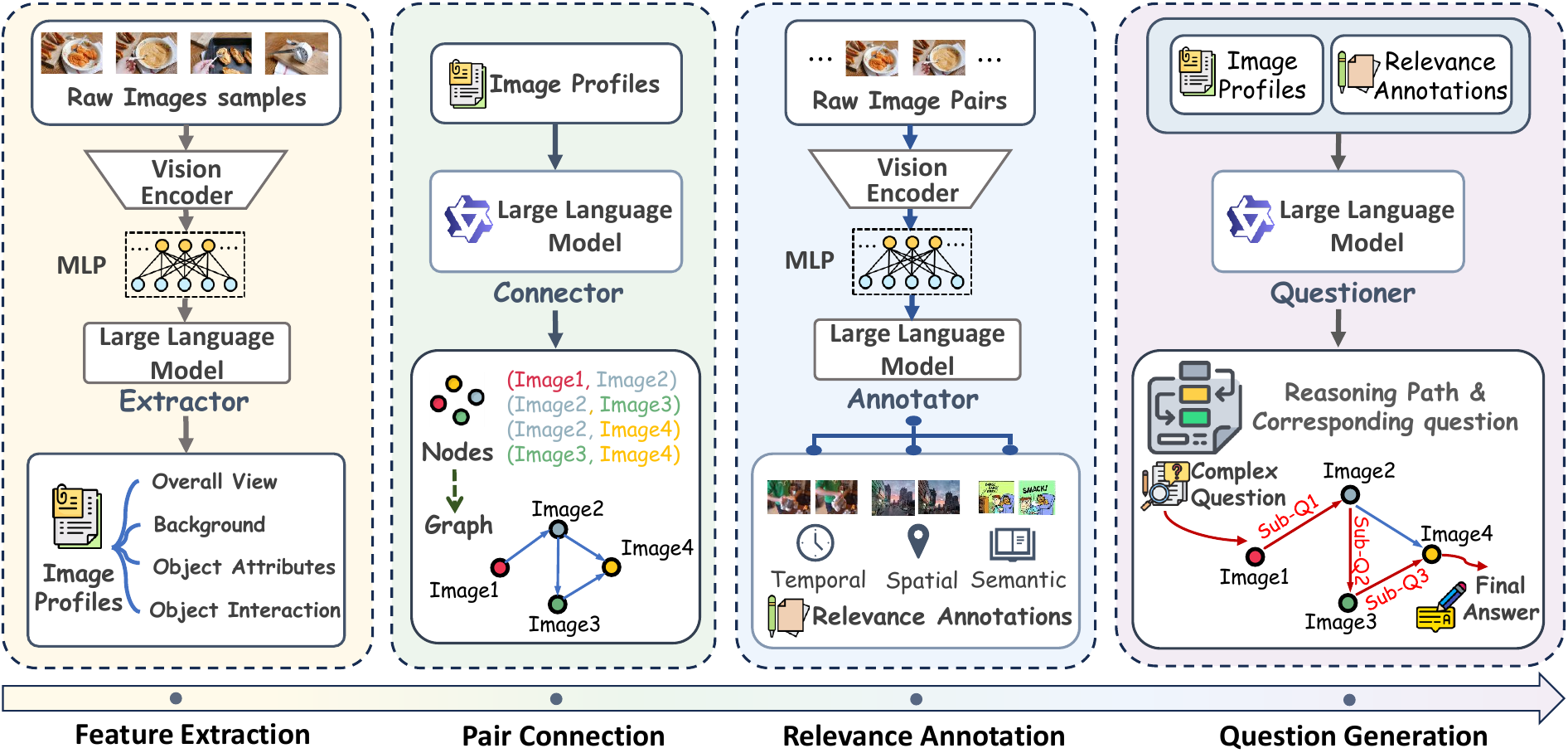}
    \caption{
    The \textit{Focus-Centric Data Synthesis} framework consists of four stages: \textbf{Feature Extraction} processes raw visual input, extracting object-level attributes and interactions into structured image profiles. \textbf{Pair Connection} links related image nodes based on their profiles. \textbf{Relevance Annotation} identifies and annotates relationships between nodes from temporal, spatial, and semantic perspectives. \textbf{Question Generation} utilizes the extracted image profiles and relationship annotations to construct multi-image reasoning paths and corresponding questions.}
    \label{pipeline}
\end{figure*}
\section{Methodology}
The methodology is detailed in two parts. First, we formulate the \textbf{\textit{Focus-Centric Visual Chain}}, which enhances the ability of VLMs to handle complex multi-image inputs through multi-step reasoning (\S~\ref{method:reasoning}). Then, we present the \textbf{\textit{Focus-Centric Data Synthesis}} framework, a bottom-up approach for synthesizing multi-image reasoning data with precise control over the reasoning process (\S~\ref{method:synthesis}).

\subsection{Focus-Centric Visual Chain}
\label{method:reasoning}

We present \textit{Focus-Centric Visual Chain}, a reasoning paradigm specifically designed for multi-image input scenarios. Given an image collection $\mathcal{G} = \{I_k | k = 1,2,\cdots,K\}$ and an initial question $Q$, the model $\mathcal{M}$ progressively constructs a reasoning chain $\mathcal{R}$ via multi-step reasoning. At each step, the model focuses on a visual evidence subset $G \subseteq \mathcal{G}$ through a dynamic selection mechanism.

Specifically, at the $i$-th reasoning step, the model $\mathcal{M}$ generates an intermediate sub-question $q_i$ and identifies its corresponding focus, a minimized visual information subset $G_i$, through: 
\begin{equation}
    q_i, G_i = \mathcal{M}(Q, \mathcal{G}, \mathcal{H}_{<i}),
\end{equation}
where $\mathcal{H}_{<i} = \{q_{1}, q_{2}, \cdots, q_{i-1}\}$ represents the sub-question history of previous steps, with $\mathcal{H}_{<1} = \emptyset$ for initialization. Subsequently, the model derives an intermediate answer $a_i$ through joint analysis of $q_i$ and $G_i$. According to the existing answer set $\mathcal{A}_{i} = \{a_{1}, a_{2}, \cdots, a_{i}\}$, the model determines whether to extend the reasoning path and outputs a stopping signal $z_i \in \{\textit{True}, \textit{False}\}$. Only when $z_i=\textit{True}$, does the model synthesize the final answer $A$ from the current QA collection $\mathcal{C}_{i} = \{(q_{1}, a_{1}), (q_{2}, a_{2}),\cdots, (q_{i}, a_{i})\}$ and terminate the reasoning; otherwise, it iteratively refocuses on distinct visual information.
Therefore, the overall reasoning process $\mathcal{R}$ can be represented as an ordered sequence:
\begin{equation}
    \mathcal{R} = \left[ (q_i, G_i, a_i, z_i) \right]_{i=1}^N,
\end{equation}
where $N$ denotes the total number of executed reasoning steps.

\subsection{Focus-Centric Data Synthesis}
\label{method:synthesis}

Aiming to implement the reasoning paradigm above in a data-driven manner, we propose \textit{Focus-Centric Data Synthesis(FCDS)}, a bottom-up evolutionary data synthesis framework. \textit{FCDS} cost-effectively facilitates the large-scale synthesis of certifiable cross-image reasoning data through open-source models.
The synthesis process begins with a set of images and incrementally assesses their interrelationships. It then formulates intermediate sub-questions, culminating in a meaningful question that aligns with a coherent reasoning path.
Specifically, \textit{FCDS} consists of four interconnected steps: \textit{Feature Extraction}, \textit{Pair connection}, \textit{Relevance Annotation} and \textit{Question Generation}, as illustrated in Figure~\ref{pipeline}. 

\paragraph{Feature Extraction.} 

Following a bottom-up manner, the synthesis process initiates with granular feature extraction, constructing a detailed profile for each image $I \in \mathcal{G}$, which consists of four core elements: (1) the overall view of $I$; (2) background descriptions; (3) object attributes and (4) object interactions. These image profiles provide a pathway for capturing object-level features and modeling their relationships.
All profiles are generated by our specialized vision-language model \textbf{Extractor}, which comprises three fundamental components: a visual encoder $f_e$ for visual feature encoding, a vision-language connector $f_c$ for modality alignment, and a large language model $f_{\phi}$ for semantic understanding as well as content generation.

Specifically, for each image $I \in \mathcal{G}$, the visual encoder $f_e$ projects $I$ into a sequence of latent embeddings:
\begin{equation}
    X^V=f_e(I)=\langle x^V_1, x^V_2, \ldots, x^V_n \rangle,
\end{equation}
where each visual token $x^V_i \in \mathbb{R}^{d^V}$ corresponds to an image patch, with $d^V$ denoting the output dimension of the visual encoder. 
The number of visual tokens $n$ depends on the visual encoder~\footnote{In some VLMs (e.g., LLaVA-OneVision), $n$ is a fixed number. While for some VLMs (e.g., Qwen2-VL), $n$ increases with the resolution of $I$.}.
Next, the vision-language connector $f_c$ performs dimension-aware projection to align visual features with textual semantics:
\begin{equation}
    X^T = f_c(X^V) = \langle x^T_1, x^T_2, \ldots, x^T_n \rangle,
\end{equation}
where each $x^T_i \in \mathbb{R}^{d^T}$ represents an aligned token. Here, $d^T$ is the dimension of $f_{\phi}$. $X^T$ is then fed into the large language model $f_{\phi}$ to generate the profile $p = f_{\phi}(X^T)$ for $I$. 
Each generated profile is treated as a node in the reasoning path.

\paragraph{Pair Connection.}
Since connecting arbitrary nodes may lack semantic validity, we establish edges only between nodes with potential relevance. 
We propose two criteria to determine node relatedness: (1) \textbf{Object-oriented}, where images share co-occurring objects, and (2) \textbf{Event-oriented}, where images depict shared or related events.
Given profile collection $\mathcal{P}$ of image set $\mathcal{G}$, a large language model \textbf{Connector} is implemented to identify valid pairwise connections:
\begin{equation}
    \{(i,j)|i,j \in [0, K),i \neq j\} = \textsf{Connector}(\mathcal{P}),
\end{equation}
where each pair $(i,j)$ indicates a potential connection between image $I_i$ and $I_j$. 
By identifying correlations between different nodes, we establish the basic structure of the reasoning path.

\paragraph{Relevance Annotation.}
To systematically characterize inter-node relevance, we classify it into three categories: Temporal, Spatial, and Semantic, as detailed as follows:

$\bullet$ \textbf{Temporal:} The paired images depict a chronological sequence, with one distinctly preceding the other in temporal succession.

$\bullet$ \textbf{Spatial:} Visual elements in paired images exhibit geometric and positional correlations, forming spatial continuity or progression.

$\bullet$ \textbf{Semantic:} The paired images exhibit intangible associations containing thematic, logical and causal relationships. Notably, this category accommodates abstract connections beyond direct visual correspondence.

To formalize the relation annotation process, we develop \textbf{Annotator}, which also comprises three components: a visual encoder $f_e$, a vision-language connector $f_c$, and a large language model $f_{\phi}$. Given connected image pair $s=(I_i,I_j)$, \textbf{Annotator} first performs independent encoding of both images followed by feature concatenation:
\begin{align}
    X^V_i&=f_e(I_i)=\langle x^V_{i,1}, x^V_{i,2}, \ldots, x^V_{i,n} \rangle, \\
    X^V_j&=f_e(I_j)=\langle x^V_{j,1}, x^V_{j,2}, \ldots, x^V_{j,n} \rangle, \\
    X^V&=concat(X^V_i,X^V_j),
\end{align}
where $X^V$ denotes the concatenated embeddings of the encoded image pair, resulting in a sequence of $2n$ visual tokens. Subsequently, $X^V$ is mapped to aligned textual tokens $X^T$ through the vision-language connector $f_c$:
\begin{align}
    X^T &= f_c(X^V) = \langle x^T_{i,1}, \ldots, x^T_{i,n}, x^T_{j,1}, \ldots, x^T_{j,n} \rangle.
\end{align}
Finally, the large language model $f_{\phi}$ generates the relation $r = f_{\phi}(X^T)$
where $r$ denotes the annotated relation for $s$. Notably, multiple relation types may be simultaneously present in a single pair.

\paragraph{Question Generation.}
The reasoning path is constructed by sampling a sequential chain of $K$ interconnected nodes. For each connected image pair $s_i$ along this path, we generate targeted sub-questions $q_i$ based on their annotated relations $r_i$ and corresponding image profiles. The question generation process is implemented through a specialized large language model \textbf{Questioner}:
\begin{equation}
    q_i = \textsf{Questioner}(s_i, r_i, p_{i,1}, p_{i,2}),
\end{equation}
where $p_{i,1}$ and $p_{i,2}$ represent the profiles of the two images in $s_i$. Finally, these sub-questions are synthesized into a coherent overarching question $Q$ through aggregation:
\begin{equation}
    Q = \textsf{Questioner}(\{q_i\}|^K_{i=1}).
\end{equation}

We utilized LLaVA-OneVision-7B~\footnote{https://huggingface.co/lmms-lab/llava-onevision-qwen2-7b-ov}~\cite{li2024llavaonevision} as the base model for the \textsf{Extractor} and \textsf{Annotator}, while Qwen2.5-7B-Instruct~\footnote{https://huggingface.co/Qwen/Qwen2.5-7B-Instruct}~\cite{qwen2.5} served as the base model for the \textsf{Connector} and \textsf{Questioner}. Through our carefully designed framework, we constructed VISC-150K, a high-quality dataset comprising 150K multi-image reasoning data instances following the \textit{Focus-Centric Visual Chain} paradigm. The image resources are collected from publicly accessible websites and include real-world photographs with diverse scenes and comics. 
More details about our dataset are introduced in Appendix~\ref{appendix:dataset}.

\begin{table*}[]
\resizebox{\textwidth}{!}{%
\begin{tabular}{lllllllll}
\toprule
\textbf{Model} & \textbf{Size} & \textbf{MMIU}  & \textbf{MuirBench} & \textbf{MIRB}  & \textbf{BLINK} & \textbf{NLVR2} & \textbf{Mantis-Eval} & \textbf{MVBench} \\ \midrule
\textcolor{gray}{GPT-4V/GPT-4o}               & -    & \textcolor{gray}{55.70} & \textcolor{gray}{68.00}     & \textcolor{gray}{53.05} & \textcolor{gray}{51.14} & \textcolor{gray}{88.80} & \textcolor{gray}{62.67}       & \textcolor{gray}{43.50}   \\
Qwen-VL-Chat                & 7B   & 15.90 & 20.42         & 14.38 & 31.17 & 58.72 & 39.17       & 42.15   \\
LLaVA-1.5                  & 7B   & 19.20 & 23.46     & 28.47 & 37.13 & 53.88 & 31.34       & 36.00   \\
LLaVA-1.6                  & 7B   & 22.20 & 27.42         & 29.83 & 39.55 & 58.88 & 45.62       & 40.90   \\
Idefics2                   & 8B   & 27.80 & 26.08     & 33.02 & 45.18 & 86.87 & 48.85       & 29.68   \\
VILA-1.5                     & 8B   & 32.45     & 33.12     & 36.52 & 39.30 & 76.45 & 51.15       & 49.40   \\
OpenFlamingo-v2             & 9B   & 22.30  & 23.73     & 28.80     & 39.18 & 36.41 & 12.44       & 7.90    \\
Mantis-Idefics2                      & 8B   & 45.60 & 44.50     & 34.82 & 49.05 & 89.71 & 57.14       & 51.38   \\
InternVL2                   & 8B   & 42.00 & 48.70      & 50.00 & 50.90 & -     & 65.40       & 65.80   \\ 
InternVL2.5                 & 8B   & 46.70 & \textbf{51.10}     & 52.50 & \underline{54.80} & -     & 67.70       & \textbf{72.00}   \\ \midrule
LLaVA-OneVision             & 7B   & 40.32 & 41.77     & 51.18 & 48.20 & 89.40 & 64.20       & 56.70   \\
\rowcolor[HTML]{E1EFE1}
\multicolumn{1}{r}{+VISC-150K}   & 7B   & 46.52$_{\textcolor{red}{(\uparrow 6.20)}}$ & \underline{49.62}$_{\textcolor{red}{(\uparrow 7.85)}}$     & 53.02$_{\textcolor{red}{(\uparrow 1.84)}}$ & 50.24$_{\textcolor{red}{(\uparrow 2.04)}}$ & \textbf{89.88}$_{\textcolor{red}{(\uparrow 0.48)}}$ & 66.36$_{\textcolor{red}{(\uparrow 2.16)}}$       &58.23$_{\textcolor{red}{(\uparrow 1.53)}}$         \\
Qwen2-VL           & 7B   & \underline{50.00} & 39.12     & \underline{58.67} & 53.20 & 86.42 & \textbf{69.60}       & 67.00   \\
\rowcolor[HTML]{E1EFE1}
\multicolumn{1}{r}{+VISC-150K} & 7B   & \textbf{52.76}$_{\textcolor{red}{(\uparrow 2.76)}}$ & 44.50$_{\textcolor{red}{(\uparrow 5.38)}}$     & \textbf{60.16}$_{\textcolor{red}{(\uparrow 1.49)}}$ & \textbf{55.34}$_{\textcolor{red}{(\uparrow 2.14)}}$ &\underline{89.82}$_{\textcolor{red}{(\uparrow 3.40)}}$   & \underline{69.12}$_{\textcolor{blue}{(\downarrow 0.48)}}$       &\underline{68.01}$_{\textcolor{red}{(\uparrow 1.01)}}$         \\ \bottomrule
\end{tabular}
}
\caption{
Performance comparison of LLaVA-OneVision and Qwen2-VL based models across seven multi-image benchmarks, with the highest scores being \textbf{bolded} and the second highest \underline{underlined}. Results highlighted in \textcolor{gray}{gray} indicate experiments using closed-source models, while the remaining results are from open-source models. 
}
\label{table:results}
\end{table*}

\section{Experiments}
\label{exp}
We first evaluate the effectiveness of our method across diverse multi-image tasks. We then conduct more investigations into our method through multifaceted experimental studies and in-depth analysis.

\subsection{Experimental Setup}
We apply \textit{Focus-Centric Visual Chains} to two pretrained models, LLaVA-OneVision-7B and Qwen2-VL-7B-Instruct\footnote{https://huggingface.co/Qwen/Qwen2-VL-7B-Instruct}~\cite{Qwen2VL}, which have been extensively fine-tuned on large-scale multi-image datasets and exhibit robust capability. 
Both models undergo LoRA~\cite{hu2022lora} fine-tuning on VISC-150k for one epoch with a batch size of 8. The learning rate is set to 1e-5 with a warmup ratio of 0.05 and as a cosine scheduler. The maximum context length is set to 32,768.

When conducting evaluations, the temperature is set to 0 and the max new tokens is 1,024. For Qwen2-VL-7B, the image resolution is cropped between 128×28×28 and 1280×28×28 to reduce memory consumption and improve inference speed. More details about experimental settings are reported in Appendix~\ref{appendix:experimental_settings}.

\subsection{Baselines}
For open-source VLMs, our baselines include Qwen2-VL~\cite{Qwen2VL}, Qwen-VL-Chat~\cite{bai2023qwen}, LLaVA-OneVision~\cite{li2024llavaonevision}, LLaVA-1.6~\cite{liu2024llavanext}, LLaVA-1.5~\cite{liu2023improved}, InternVL2.5~\cite{chen2024expanding}, InternVL2~\cite{chen2024far}, Mantis-Idefics2~\cite{jiang2024mantis}, Idefics2~\cite{laurencon2024idefics2}, VILA-1.5~\cite{lin2023vila} and OpenFlamingo-v2~\cite{awadalla2023openflamingo}. Among close-source VLMs, we select GPT-4V/GPT-4o~\cite{openai2024gpt4o} as the baseline. Please refer to Appendix~\ref{appendix:baselines} for more details.

\subsection{Benchmarks}

We evaluate our method on seven comprehensive multi-image benchmarks that span diverse scenarios.
The statics of each benchmark are detailed in Appendix~\ref{appendix:benchmarks}, with characteristics listed as follows:

\noindent(1) \textbf{MMIU}~\cite{meng2024mmiu} categorizes multi-image relationships into three primary types: semantic, spatial, and temporal. These categories are further subdivided into seven subtypes, covering 52 distinct multi-image understanding tasks. 

\noindent(2) \textbf{MuirBench}~\cite{wang2024muirbench} covers 12 distinct multi-image understanding tasks and encompassing 10 types of multi-image relationships. 

\noindent(3) \textbf{MIRB}~\cite{zhao2024benchmarking}
includes four evaluation dimensions: perception, visual world knowledge, reasoning, and multi-hop reasoning. Each category consists of tasks requiring comparison and inference across multiple images.

\noindent(4) \textbf{BLINK}~\cite{fu2024blink} incorporates 14 visual perception tasks that humans can quickly solve, covering indoor, outdoor, and natural scenes.

\noindent(5) \textbf{NLVR2}~\cite{suhr2019nlvr2} 
contains examples of English sentences paired with online photos, focusing on rich linguistic and visual content to support diverse reasoning tasks.

\noindent(6) \textbf{Mantis-Eval}~\cite{jiang2024mantis} comprises 
high-quality multi-image reasoning samples, designed for diverse multi-image skills such as co-reference, reasoning, and comparison.

\noindent(7) \textbf{MVBench}~\cite{li2024mvbench} consists of 20 challenging video understanding tasks,
which cover a wide range of temporal understanding skills in video scenarios.

\subsection{Results}
The experimental results are presented in Table~\ref{table:results}. Both LLaVA-OneVision and Qwen2-VL demonstrate consistent performance improvements across seven benchmarks after fine-tuning with our synthesized dataset VISC-150K, indicating the effectiveness of our method across different VLM architectures.

LLaVA-OneVision achieves benchmark-leading improvements of 6.20\% on MMIU and 7.85\% on MuirBench, demonstrating breakthrough capabilities. Significantly, our method elevates even the already superior Qwen2-VL model across multiple benchmarks, achieving an average gain of 2.24\% over its strong baselines. When combined with LLaVA-OneVision's 3.16\% average improvement, these consistent enhancements across both high-performing and emerging models conclusively validate the universality of our approach.

Our method establishes new state-of-the-art results on four benchmarks: MMIU, MIRB, BLINK, and NLVR2.
These improvements can be attributed to three key characteristics: (1) the richness of visual information, (2) the diversity of inter-image relationships, and (3) the complexity of task formulations. These characteristics align well with our method's enhanced capabilities in visual perception, comprehension, and reasoning.

Moreover, our method demonstrates measurable performance gains on the video benchmark MVBench. Given that a video is essentially a collection of frames, it falls into a multi-image scenario with temporal correlations, where our method maintains efficacy. This result substantiates that our \textit{Focus-Centric Visual Chain} reasoning paradigm operates in a domain-agnostic manner, exhibiting robust applicability as well as transferability.

\subsection{Discussions}
In this section, we propose five research questions and conduct an in-depth investigation to provide a more comprehensive evaluation of our method.

\begin{figure}[t]
    \centering
    \includegraphics[width=\columnwidth]{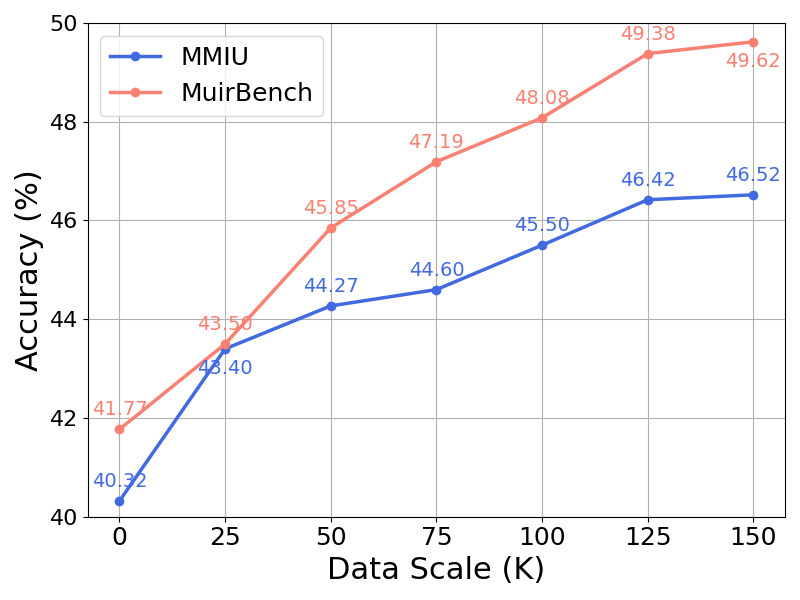}
    \caption{
    The impact of dataset scale on LLaVA-OneVision's performance across MMIU and MuirBench benchmarks. As the data scale increases, the model's accuracy progressively improves.
    }
    \label{line plot}
\end{figure}

\noindent\textbf{RQ1: How does data size affect performance?}

To investigate the impact of the data scale, we create five subsets from VISC-150K through random sampling, which contain 25K, 50K, 75K, 100K, and 125K instances respectively. Each subset is used to fine-tune LLaVA-OneVision using LoRA for one epoch. The fine-tuned models are then evaluated on the MMIU and MuirBench, and the result is visualized in Figure~\ref{line plot}.

The performance curves reveal a non-linear relationship: the model achieves rapid improvements when increasing data size from 0 to 125K, followed by a more gradual improvement from 125K to 150K, suggesting diminishing returns but continued learning potential. 
We attribute the rapid performance leap observed with the 0–25K data scale to a capability activation process, where the model unlocks stronger multi-image potential by learning from reasoning data under the new paradigm.  
This result suggests that data constructed through \textit{Focus-Centric Data Synthesis} framework can be effectively scaled up, which is essential for further enhancing performance by expanding the data size.

\noindent\textbf{RQ2: What is the effect on different sub-tasks?}

\begin{figure}[t]
    \centering
    \includegraphics[width=\columnwidth]{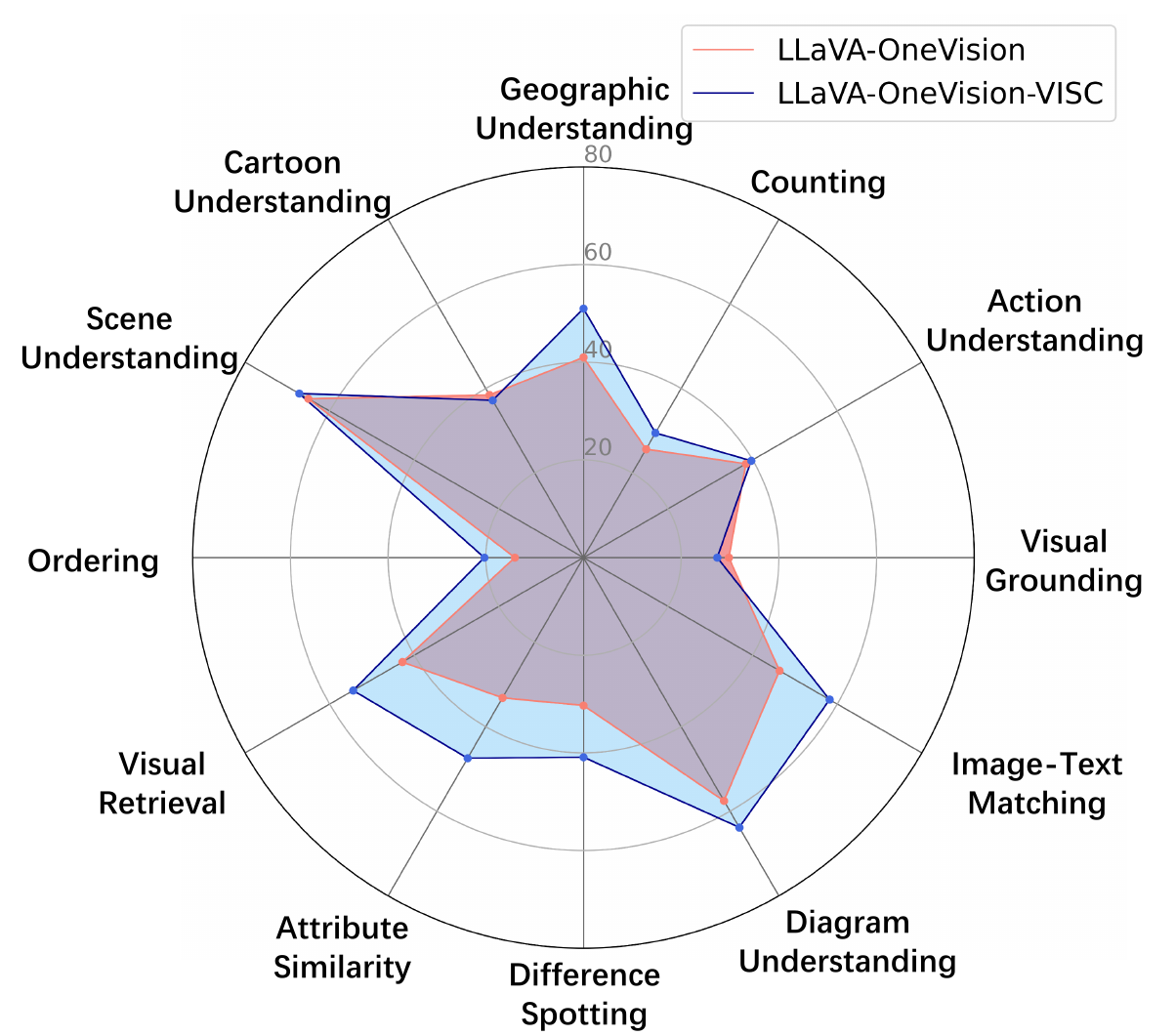}
    \caption{The accuracy comparison of LLaVA-OneVision on 12 MuirBench sub-tasks with and without being fine-tuned on VISC-150K.}
    \label{Radar chart}
\end{figure}

To better understand our approach's strengths, we conducted a detailed analysis across different multi-image tasks in MuirBench. 
We compare the performance of LLaVA-OneVision before and after being finetuned on VISC-150K, as illustrated in Figure~\ref{Radar chart}. 
Across all twelve sub-tasks in MuirBench, only four of them do not exhibit substantial improvements. These tasks face dual constraints stemming from (1) intrinsic limitations in vision-language architectures and (2) capacity constraints of foundational language models. This is exemplified by the 3D spatial reasoning required for \textit{Scene Understanding} and the nuanced semantic interpretation essential for \textit{Cartoon Understanding}. 

In contrast, the remaining eight tasks show significant improvements. These tasks predominantly involve similarity analysis or comparative reasoning at the image or feature level, such as visual retrieval and attribute similarity. 
Additionally, some of these sub-tasks (e.g., geographic understanding) involve image types and task categories that are not present in VISC-150K, which emphasize the cross-domain generalization capabilities of our method.

\begin{figure}[t]
    \centering
    \includegraphics[width=\columnwidth]{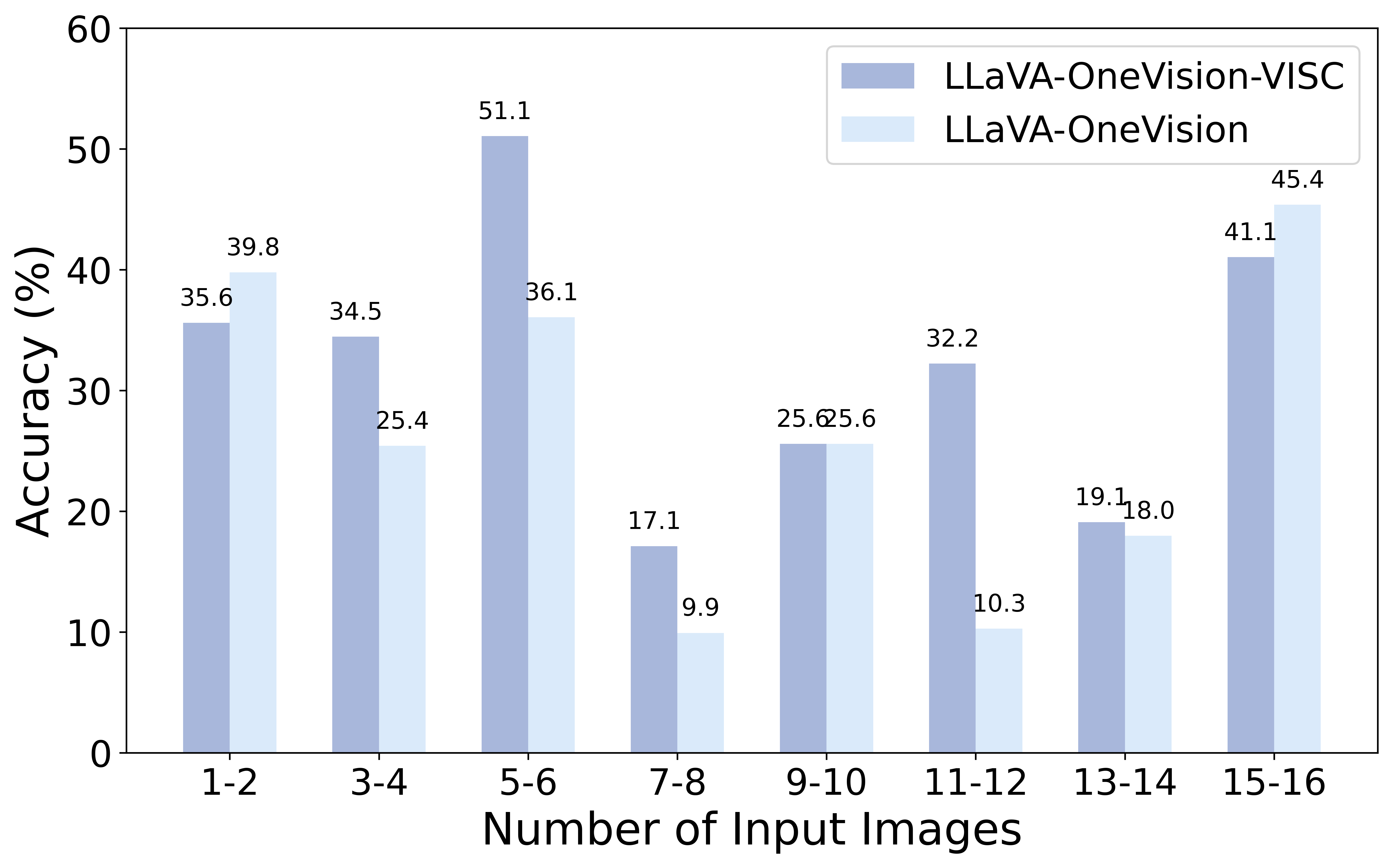}
    \caption{The distribution of task accuracy for LLaVA-OneVision based models across varying numbers of input images, grouped into eight buckets from 1 to 16 images with an interval of 2.}
    \label{bucket analysis}
\end{figure}

\noindent\textbf{RQ3: How does the number of input images impact performance?}

To investigate how the number of input images impacts our method, 
we conducted a detailed analysis of LLaVA-OneVision on the MMIU benchmark.
The instances are grouped into different buckets by the number of input images.

As demonstrated in Figure~\ref{bucket analysis},  
when handling 3-8 images, LLaVA-OneVision-VISC achieves remarkable improvements, suggesting successful identification of cross-image relationships in medium-sized image sets. 
This capability persists even with larger inputs (11-14 images), where LLaVA-OneVision-VISC maintains superior performance compared to baseline models while avoiding performance degradation from information overload.  
However, when processing more than 15 images, the performance of LLaVA-OneVision-VISC exhibits slight degradation, which may be attributed to amplified noise levels or interference from irrelevant data patterns in prolonged image sequences.

\noindent\textbf{RQ4: Does VISC-150K affect general ability?}

Although our method demonstrates remarkable improvements in multi-image tasks, it is crucial to evaluate whether these gains come at the expense of general task performance. To investigate this issue, we adopt Qwen2-VL as the base model and conduct analysis on four benchmarks in other domains. 
These benchmarks are based on single-image inputs and evaluate the fine-tuned model's performance from various perspectives, including hallucination, single-image data language capability, domain-specific knowledge, and mathematical reasoning. Specifically: \textbf{HallusionBench}~\cite{guan2023hallusionbench} is designed to assess VLMs' ability to comprehend and interpret visual data; \textbf{MMStar}~\cite{chen2024we} requires advanced multi-modal capabilities for accurate interpretation; \textbf{MMMU}~\cite{yue2024mmmu} focuses on evaluating models' ability to apply domain-specific knowledge; \textbf{MathVista}~\cite{lu2023mathvista} integrates mathematical reasoning with visual tasks.
The results are presented in Table~\ref{table:vertical_exp}.
The model fine-tuned on VISC-150K maintains comparable or superior performance across all benchmarks, 
indicating that our method enhances the model's ability to perceive visual information without diminishing its general vision-language capabilities.

\begin{table}[t]
\resizebox{\columnwidth}{!}{%
\begin{tabular}{@{}lcccc@{}}
\toprule
\textbf{Model} & \textbf{HallusionBench} & \textbf{MMStar} & \textbf{MMMU} & \textbf{MathVista} \\ \midrule
Qwen2-VL-7B      & 64.8           & 60.7 & 54.1 & 58.2      \\
Qwen2-VL-7B-VISC & 66.3               &60.4      &54.3      &58.5           \\ \bottomrule
\end{tabular}%
}

\caption{After fine-tuning, the model based on the Qwen2-VL architecture shows comparable or slightly improved performance across four vertical vision-language benchmarks.}
\label{table:vertical_exp}
\end{table}

\noindent\textbf{RQ5: How is the quality of synthesized data?}

Quality assurance is particularly critical for our fully automated data synthesis process. 
We conducted a rigorous quality assessment by engaging three Ph.D. students with expertise in computer vision and natural language processing. 
They evaluated a stratified random sample of 200 instances, with each instance independently assessed by all annotators. The overall accuracy is calculated based on three metrics: focus accuracy, sub-question correctness, and final answer correctness. A sample is counted as a positive instance when all three metrics are unanimously assessed as correct by at least two annotators, or regarded as a negative instance. With the synthesized data attaining 97.5\% overall accuracy (Fleiss' $\kappa$~\cite{fleiss1971measuring} = 0.637), these results robustly validate the method's reliability across measurement dimensions.

\section{Conclusion}

This study addresses the challenges from multi-image scenarios through two key innovations: (1) the \textit{Focus-Centric Visual Chain}, a structured reasoning paradigm that breaks down complex tasks into targeted sub-questions with explicit visual focus, and (2) a bottom-up framework for automated synthesis of reasoning data in the form of \textit{Focus-Centric Visual Chain}. Through this methodology, a large-scale multimodal dataset of 150K instances featuring multi-image compositions and cascaded reasoning chains, systematically generated via our automated synthesis framework. 
Models fine-tuned on VISC-150K achieve consistent improvements across multiple multi-image benchmarks. Our work not only breaks performance ceilings in existing tasks but also establishes an advanced framework for data-driven visual reasoning, providing actionable pathways to resolve persistent bottlenecks on multi-image tasks.

\section*{Limitations}

While our approach demonstrates promising results, we identify several important limitations. The \textit{Focus-Centric Data Synthesis} framework requires pairwise relevance annotation across images, leading to quadratic computational complexity. We maintain moderate sizes of image sets to balance data diversity and computational efficiency.

Moreover, Our VISC-150K dataset primarily focuses on real-world photographs and comics. The approach's effectiveness remains untested on structured visual content such as charts, diagrams, and code snippets, which may require different reasoning patterns.

In addition, the current implementation is constrained by the language models' inherent capabilities. Consequently, our method inherits their limitations in managing complex spatial dynamics, domain-exclusive contexts, and expertise-dependent visual subtleties.

These limitations suggest promising directions for future research in multi-image understanding and reasoning.

\bibliography{custom}

\begin{thebibliography}{51}
\providecommand{\natexlab}[1]{#1}

\bibitem[{Alayrac et~al.(2022)Alayrac, Donahue, Luc, Miech, Barr, Hasson, Lenc, Mensch, Millican, Reynolds, Ring, Rutherford, Cabi, Han, Gong, Samangooei, Monteiro, Menick, Borgeaud, Brock, Nematzadeh, Sharifzadeh, Binkowski, Barreira, Vinyals, Zisserman, and Simonyan}]{alayrac2022flamingovisuallanguagemodel}
Jean-Baptiste Alayrac, Jeff Donahue, Pauline Luc, Antoine Miech, Iain Barr, Yana Hasson, Karel Lenc, Arthur Mensch, Katie Millican, Malcolm Reynolds, Roman Ring, Eliza Rutherford, Serkan Cabi, Tengda Han, Zhitao Gong, Sina Samangooei, Marianne Monteiro, Jacob Menick, Sebastian Borgeaud, Andrew Brock, Aida Nematzadeh, Sahand Sharifzadeh, Mikolaj Binkowski, Ricardo Barreira, Oriol Vinyals, Andrew Zisserman, and Karen Simonyan. 2022.
\newblock \href {https://arxiv.org/abs/2204.14198} {Flamingo: a visual language model for few-shot learning}.
\newblock \emph{Preprint}, arXiv:2204.14198.

\bibitem[{Awadalla et~al.(2023)Awadalla, Gao, Gardner, Hessel, Hanafy, Zhu, Marathe, Bitton, Gadre, Sagawa, Jitsev, Kornblith, Koh, Ilharco, Wortsman, and Schmidt}]{awadalla2023openflamingo}
Anas Awadalla, Irena Gao, Josh Gardner, Jack Hessel, Yusuf Hanafy, Wanrong Zhu, Kalyani Marathe, Yonatan Bitton, Samir Gadre, Shiori Sagawa, Jenia Jitsev, Simon Kornblith, Pang~Wei Koh, Gabriel Ilharco, Mitchell Wortsman, and Ludwig Schmidt. 2023.
\newblock Openflamingo: An open-source framework for training large autoregressive vision-language models.
\newblock \emph{arXiv preprint arXiv:2308.01390}.

\bibitem[{Bai et~al.(2023)Bai, Bai, Yang, Wang, Tan, Wang, Lin, Zhou, and Zhou}]{bai2023qwen}
Jinze Bai, Shuai Bai, Shusheng Yang, Shijie Wang, Sinan Tan, Peng Wang, Junyang Lin, Chang Zhou, and Jingren Zhou. 2023.
\newblock \href {https://arxiv.org/abs/2308.12966} {Qwen-vl: A versatile vision-language model for understanding, localization, text reading, and beyond}.
\newblock \emph{arXiv preprint arXiv:2308.12966}.

\bibitem[{Besta et~al.(2023)Besta, Blach, Kubicek, Gerstenberger, Gianinazzi, Gajda, Lehmann, Podstawski, Niewiadomski, Nyczyk, and Hoefler}]{besta2023graph}
Maciej Besta, Nils Blach, Ales Kubicek, Robert Gerstenberger, Lukas Gianinazzi, Joanna Gajda, Tomasz Lehmann, Michal Podstawski, Hubert Niewiadomski, Piotr Nyczyk, and Torsten Hoefler. 2023.
\newblock \href {https://arxiv.org/abs/2308.09687} {Graph of thoughts: Solving elaborate problems with large language models}.
\newblock \emph{arXiv preprint arXiv:2308.09687}.

\bibitem[{Campbell et~al.(2024)Campbell, Rane, Giallanza, Sabbata, Ghods, Joshi, Ku, Frankland, Griffiths, Cohen, and Webb}]{campbell2024understanding}
Declan~Iain Campbell, Sunayana Rane, Tyler Giallanza, C.~Nicolò~De Sabbata, Kia Ghods, Amogh Joshi, Alexander Ku, Steven~M. Frankland, Thomas~L. Griffiths, Jonathan~D. Cohen, and Taylor~Whittington Webb. 2024.
\newblock \href {https://arxiv.org/abs/2411.00238} {Understanding the limits of vision language models through the lens of the binding problem}.
\newblock \emph{arXiv preprint arXiv:2411.00238}.

\bibitem[{Chen et~al.(2024{\natexlab{a}})Chen, Li, Dong, Zhang, Zang, Chen, Duan, Wang, Qiao, Lin et~al.}]{chen2024we}
Lin Chen, Jinsong Li, Xiaoyi Dong, Pan Zhang, Yuhang Zang, Zehui Chen, Haodong Duan, Jiaqi Wang, Yu~Qiao, Dahua Lin, et~al. 2024{\natexlab{a}}.
\newblock Are we on the right way for evaluating large vision-language models?
\newblock \emph{arXiv preprint arXiv:2403.20330}.

\bibitem[{Chen et~al.(2024{\natexlab{b}})Chen, Wang, Tian, Ye, Gao, Cui, Tong, Hu, Luo, Ma et~al.}]{chen2024expanding}
Zhe Chen, Weiyun Wang, Hao Tian, Shenglong Ye, Zhangwei Gao, Erfei Cui, Wenwen Tong, Kongzhi Hu, Jiapeng Luo, Zheng Ma, et~al. 2024{\natexlab{b}}.
\newblock \href {https://arxiv.org/abs/2412.05271} {Expanding performance boundaries of open-source multimodal models with model, data, and test-time scaling}.
\newblock \emph{arXiv preprint arXiv:2412.05271}.

\bibitem[{Chen et~al.(2024{\natexlab{c}})Chen, Wang, Tian, Ye, Gao, Cui, Tong, Hu, Luo, Ma et~al.}]{chen2024far}
Zhe Chen, Weiyun Wang, Hao Tian, Shenglong Ye, Zhangwei Gao, Erfei Cui, Wenwen Tong, Kongzhi Hu, Jiapeng Luo, Zheng Ma, et~al. 2024{\natexlab{c}}.
\newblock \href {https://arxiv.org/abs/2404.16821} {How far are we to gpt-4v? closing the gap to commercial multimodal models with open-source suites}.
\newblock \emph{arXiv preprint arXiv:2404.16821}.

\bibitem[{Cheng et~al.(2024)Cheng, Guan, Wu, and Yan}]{cheng2024least}
Chuanqi Cheng, Jian Guan, Wei Wu, and Rui Yan. 2024.
\newblock From the least to the most: Building a plug-and-play visual reasoner via data synthesis.
\newblock \emph{arXiv preprint arXiv:2406.19934}.

\bibitem[{Dai et~al.(2023)Dai, Li, Li, Tiong, Zhao, Wang, Li, Fung, and Hoi}]{dai2023instructblip}
Wenliang Dai, Junnan Li, Dongxu Li, Anthony Meng~Huat Tiong, Junqi Zhao, Weisheng Wang, Boyang Li, Pascale Fung, and Steven Hoi. 2023.
\newblock \href {https://arxiv.org/abs/2305.06500} {{InstructBLIP}: Towards general-purpose vision-language models with instruction tuning}.
\newblock \emph{arXiv preprint arXiv:2305.06500}.

\bibitem[{Daniali and Kim(2023)}]{daniali2023perception}
Maryam Daniali and Edward Kim. 2023.
\newblock Perception over time: Temporal dynamics for robust image understanding.
\newblock In \emph{Proceedings of the IEEE/CVF Conference on Computer Vision and Pattern Recognition}, pages 5656--5665.

\bibitem[{Fleiss(1971)}]{fleiss1971measuring}
Joseph~L. Fleiss. 1971.
\newblock \href {https://doi.org/10.1037/h0031619} {Measuring nominal scale agreement among many raters}.
\newblock \emph{Psychological Bulletin}, 76(5):378--382.

\bibitem[{Fu et~al.(2024)Fu, Hu, Li, Feng, Wang, Lin, Roth, Smith, Ma, and Krishna}]{fu2024blink}
Xingyu Fu, Yushi Hu, Bangzheng Li, Yu~Feng, Haoyu Wang, Xudong Lin, Dan Roth, Noah~A Smith, Wei-Chiu Ma, and Ranjay Krishna. 2024.
\newblock Blink: Multimodal large language models can see but not perceive.
\newblock In \emph{European Conference on Computer Vision}, pages 148--166. Springer.

\bibitem[{Gao et~al.(2023)Gao, Pi, Zhang, Ye, Zhong, Wang, Hong, Han, Xu, Li, and Kong}]{gao2023gllava}
Jiahui Gao, Renjie Pi, Jipeng Zhang, Jiacheng Ye, Wanjun Zhong, Yufei Wang, Lanqing Hong, Jianhua Han, Hang Xu, Zhenguo Li, and Lingpeng Kong. 2023.
\newblock \href {https://arxiv.org/abs/2312.11370} {{G-LLaVA: Solving Geometric Problem with Multi-Modal Large Language Model}}.
\newblock \emph{arXiv preprint arXiv:2312.11370}.

\bibitem[{Gemini~Team(2024)}]{gemini2024}
Google Gemini~Team. 2024.
\newblock \href {https://arxiv.org/abs/2403.05530} {Gemini 1.5: Unlocking multimodal understanding across millions of tokens of context}.
\newblock \emph{arXiv preprint arXiv:2403.05530}.

\bibitem[{Guan et~al.(2023)Guan, Liu, Wu, Xian, Li, Liu, Wang, Chen, Huang, Yacoob et~al.}]{guan2023hallusionbench}
Tianrui Guan, Fuxiao Liu, Xiyang Wu, Ruiqi Xian, Zongxia Li, Xiaoyu Liu, Xijun Wang, Lichang Chen, Furong Huang, Yaser Yacoob, et~al. 2023.
\newblock Hallusionbench: An advanced diagnostic suite for entangled language hallucination and visual illusion in large vision-language models.
\newblock \emph{arXiv preprint arXiv:2310.14566}.

\bibitem[{Hu et~al.(2022)Hu, Shen, Wallis, Allen-Zhu, Li, Wang, Wang, and Chen}]{hu2022lora}
Edward~J. Hu, Yelong Shen, Phillip Wallis, Zeyuan Allen-Zhu, Yuanzhi Li, Shean Wang, Lu~Wang, and Weizhu Chen. 2022.
\newblock \href {https://openreview.net/forum?id=nZeVKeeFYf9} {{LoRA}: Low-rank adaptation of large language models}.
\newblock In \emph{International Conference on Learning Representations (ICLR)}.

\bibitem[{Huang et~al.(2023)Huang, Dong, Wang, Hao, Singhal, Ma, Lv, Cui, Mohammed, Patra, Liu, Aggarwal, Chi, Bjorck, Chaudhary, Som, Song, and Wei}]{huang2023languageneedaligningperception}
Shaohan Huang, Li~Dong, Wenhui Wang, Yaru Hao, Saksham Singhal, Shuming Ma, Tengchao Lv, Lei Cui, Owais~Khan Mohammed, Barun Patra, Qiang Liu, Kriti Aggarwal, Zewen Chi, Johan Bjorck, Vishrav Chaudhary, Subhojit Som, Xia Song, and Furu Wei. 2023.
\newblock \href {https://arxiv.org/abs/2302.14045} {Language is not all you need: Aligning perception with language models}.
\newblock \emph{Preprint}, arXiv:2302.14045.

\bibitem[{Jiang et~al.(2024)Jiang, He, Zeng, Wei, Ku, Liu, and Chen}]{jiang2024mantis}
Dongfu Jiang, Xuan He, Huaye Zeng, Cong Wei, Max Ku, Qian Liu, and Wenhu Chen. 2024.
\newblock \href {https://arxiv.org/abs/2405.01483} {{MANTIS}: Interleaved multi-image instruction tuning}.
\newblock \emph{arXiv preprint arXiv:2405.01483}.

\bibitem[{Laurençon et~al.(2024)Laurençon, Tronchon, Cord, and Sanh}]{laurencon2024idefics2}
Hugo Laurençon, Léo Tronchon, Matthieu Cord, and Victor Sanh. 2024.
\newblock \href {https://arxiv.org/abs/2405.02246} {What matters when building vision-language models?}
\newblock \emph{arXiv preprint arXiv:2405.02246}.

\bibitem[{Lee et~al.(2024)Lee, Wang, Li, and Zhang}]{lee2024multimodal}
Junlin Lee, Yequan Wang, Jing Li, and Min Zhang. 2024.
\newblock Multimodal reasoning with multimodal knowledge graph.
\newblock \emph{arXiv preprint arXiv:2406.02030}.

\bibitem[{Li et~al.(2024{\natexlab{a}})Li, Zhang, Guo, Zhang, Li, Zhang, Zhang, Zhang, Li, Liu, and Li}]{li2024llavaonevision}
Bo~Li, Yuanhan Zhang, Dong Guo, Renrui Zhang, Feng Li, Hao Zhang, Kaichen Zhang, Peiyuan Zhang, Yanwei Li, Ziwei Liu, and Chunyuan Li. 2024{\natexlab{a}}.
\newblock \href {https://arxiv.org/abs/2408.03326} {{LLaVA-OneVision}: Easy visual task transfer}.
\newblock \emph{arXiv preprint arXiv:2408.03326}.

\bibitem[{Li et~al.(2023)Li, Li, Savarese, and Hoi}]{li2023blip2}
Junnan Li, Dongxu Li, Silvio Savarese, and Steven Hoi. 2023.
\newblock {BLIP}-2: Bootstrapping language-image pre-training with frozen image encoders and large language models.
\newblock In \emph{Proceedings of the 40th International Conference on Machine Learning}, pages 19730--19742. PMLR.

\bibitem[{Li et~al.(2024{\natexlab{b}})Li, Zhang, Zhang, Zhang, Zhang, Wang, Zhang, Wang, Li, Zhang et~al.}]{li2024mvbench}
Yining Li, Yizhuo Zhang, Yifan Zhang, Zihan Zhang, Yuchen Zhang, Yixin Wang, Yixin Zhang, Yizhuo Wang, Yining Li, Yizhuo Zhang, et~al. 2024{\natexlab{b}}.
\newblock \href {https://arxiv.org/abs/2406.12345} {{MV-Bench: A Comprehensive Benchmark for Multimodal Video Understanding}}.
\newblock \emph{arXiv preprint arXiv:2406.12345}.

\bibitem[{Lin et~al.(2023)Lin, Yin, Zhang, Chen, Yuan, and Han}]{lin2023vila}
Ji~Lin, Hongxu Yin, Chunyuan Zhang, Xiyang Chen, Lu~Yuan, and Song Han. 2023.
\newblock \href {https://arxiv.org/abs/2312.07533} {{VILA: On Pre-training for Visual Language Models}}.
\newblock \emph{arXiv preprint arXiv:2312.07533}.

\bibitem[{Liu et~al.(2023{\natexlab{a}})Liu, Li, Li, and Lee}]{liu2023improved}
Haotian Liu, Chunyuan Li, Yuheng Li, and Yong~Jae Lee. 2023{\natexlab{a}}.
\newblock \href {https://arxiv.org/abs/2310.03744} {Improved baselines with visual instruction tuning}.
\newblock \emph{arXiv preprint arXiv:2310.03744}.

\bibitem[{Liu et~al.(2024{\natexlab{a}})Liu, Li, Li, Li, Zhang, Shen, and Lee}]{liu2024llavanext}
Haotian Liu, Chunyuan Li, Yuheng Li, Bo~Li, Yuanhan Zhang, Sheng Shen, and Yong~Jae Lee. 2024{\natexlab{a}}.
\newblock \href {https://llava-vl.github.io/blog/2024-01-30-llava-next/} {{LLaVA-NeXT: Improved reasoning, OCR, and world knowledge}}.

\bibitem[{Liu et~al.(2023{\natexlab{b}})Liu, Li, Wu, and Lee}]{liu2023visual}
Haotian Liu, Chunyuan Li, Qingyang Wu, and Yong~Jae Lee. 2023{\natexlab{b}}.
\newblock \href {https://arxiv.org/abs/2304.08485} {Visual instruction tuning}.
\newblock \emph{arXiv preprint arXiv:2304.08485}.

\bibitem[{Liu et~al.(2024{\natexlab{b}})Liu, Chen, Zhang, Gao, Zhang, and Yan}]{liu2024skepticism}
Yuhan Liu, Xiuying Chen, Xiaoqing Zhang, Xing Gao, Ji~Zhang, and Rui Yan. 2024{\natexlab{b}}.
\newblock From skepticism to acceptance: simulating the attitude dynamics toward fake news.
\newblock In \emph{Proceedings of the Thirty-Third International Joint Conference on Artificial Intelligence}, pages 7886--7894.

\bibitem[{Liu et~al.(2024{\natexlab{c}})Liu, Song, Zhang, Chen, and Yan}]{liu2024tiny}
Yuhan Liu, Zirui Song, Xiaoqing Zhang, Xiuying Chen, and Rui Yan. 2024{\natexlab{c}}.
\newblock From a tiny slip to a giant leap: An llm-based simulation for fake news evolution.
\newblock \emph{arXiv preprint arXiv:2410.19064}.

\bibitem[{Lu et~al.(2023)Lu, Bansal, Xia, Liu, Li, Hajishirzi, Cheng, Chang, Galley, and Gao}]{lu2023mathvista}
Pan Lu, Hritik Bansal, Tony Xia, Jiacheng Liu, Chunyuan Li, Hannaneh Hajishirzi, Hao Cheng, Kai-Wei Chang, Michel Galley, and Jianfeng Gao. 2023.
\newblock Mathvista: Evaluating mathematical reasoning of foundation models in visual contexts.
\newblock \emph{arXiv preprint arXiv:2310.02255}.

\bibitem[{Meng et~al.(2024)Meng, Wang, Li, Lu, Tian, Liao, Zhu, Dai, Qiao, Luo, Zhang, and Shao}]{meng2024mmiu}
Fanqing Meng, Jin Wang, Chuanhao Li, Quanfeng Lu, Hao Tian, Jiaqi Liao, Xizhou Zhu, Jifeng Dai, Yu~Qiao, Ping Luo, Kaipeng Zhang, and Wenqi Shao. 2024.
\newblock \href {https://arxiv.org/abs/2408.02718} {{MMIU}: Multimodal multi-image understanding for evaluating large vision-language models}.
\newblock \emph{arXiv preprint arXiv:2408.02718}.

\bibitem[{OpenAI(2024{\natexlab{a}})}]{openai2024gpt4o}
OpenAI. 2024{\natexlab{a}}.
\newblock \href {https://arxiv.org/abs/2410.21276} {{GPT-4o} system card}.
\newblock \emph{arXiv preprint arXiv:2410.21276}.

\bibitem[{OpenAI(2024{\natexlab{b}})}]{openai2024o1}
OpenAI. 2024{\natexlab{b}}.
\newblock \href {https://openai.com/index/openai-o1-system-card/} {{OpenAI o1 System Card}}.

\bibitem[{Peng et~al.(2023)Peng, Wang, Dong, Hao, Huang, Ma, and Wei}]{peng2023kosmos2groundingmultimodallarge}
Zhiliang Peng, Wenhui Wang, Li~Dong, Yaru Hao, Shaohan Huang, Shuming Ma, and Furu Wei. 2023.
\newblock \href {https://arxiv.org/abs/2306.14824} {Kosmos-2: Grounding multimodal large language models to the world}.
\newblock \emph{Preprint}, arXiv:2306.14824.

\bibitem[{Shi et~al.(2024)Shi, Li, Zhang, Zhang, Zhang, Wang, Li, Zhang, Wang, Zhang et~al.}]{shi2024mathllava}
Yichong Shi, Yucheng Li, Yifan Zhang, Yizhuo Zhang, Yuchen Zhang, Yixin Wang, Yining Li, Zihan Zhang, Yizhuo Wang, Yixin Zhang, et~al. 2024.
\newblock \href {https://aclanthology.org/2024.findings-emnlp.268} {{Math-LLaVA: Bootstrapping Mathematical Reasoning for Multimodal Large Language Models}}.
\newblock In \emph{Findings of the Association for Computational Linguistics: EMNLP 2024}.

\bibitem[{Suhr et~al.(2019)Suhr, Zhou, Zhang, Zhang, Bai, Xiong, and Artzi}]{suhr2019nlvr2}
Alane Suhr, Stephanie Zhou, Iris Zhang, Huajun Zhang, Rui Bai, Yichen Xiong, and Yoav Artzi. 2019.
\newblock \href {https://aclanthology.org/P19-1644/} {{A Corpus for Reasoning about Natural Language Grounded in Photographs}}.
\newblock In \emph{Proceedings of the 57th Annual Meeting of the Association for Computational Linguistics}, pages 6418--6428.

\bibitem[{Wang et~al.(2024{\natexlab{a}})Wang, Fu, Huang, Li, Liu, Liu, Ma, Xu, Zhou, Zhang et~al.}]{wang2024muirbench}
Fei Wang, Xingyu Fu, James~Y Huang, Zekun Li, Qin Liu, Xiaogeng Liu, Mingyu~Derek Ma, Nan Xu, Wenxuan Zhou, Kai Zhang, et~al. 2024{\natexlab{a}}.
\newblock \href {https://arxiv.org/abs/2406.09411} {Muirbench: A comprehensive benchmark for robust multi-image understanding}.
\newblock \emph{arXiv preprint arXiv:2406.09411}.

\bibitem[{Wang et~al.(2024{\natexlab{b}})Wang, Bai, Tan, Wang, Fan, Bai, Chen, Liu, Wang, Ge, Fan, Dang, Du, Ren, Men, Liu, Zhou, Zhou, and Lin}]{Qwen2VL}
Peng Wang, Shuai Bai, Sinan Tan, Shijie Wang, Zhihao Fan, Jinze Bai, Keqin Chen, Xuejing Liu, Jialin Wang, Wenbin Ge, Yang Fan, Kai Dang, Mengfei Du, Xuancheng Ren, Rui Men, Dayiheng Liu, Chang Zhou, Jingren Zhou, and Junyang Lin. 2024{\natexlab{b}}.
\newblock Qwen2-vl: Enhancing vision-language model's perception of the world at any resolution.
\newblock \emph{arXiv preprint arXiv:2409.12191}.

\bibitem[{Wang et~al.(2023)Wang, Wei, Schuurmans, Le, Chi, Narang, Chowdhery, and Zhou}]{wang2023self}
Xuezhi Wang, Jason Wei, Dale Schuurmans, Quoc Le, Ed~Chi, Sharan Narang, Aakanksha Chowdhery, and Denny Zhou. 2023.
\newblock \href {https://openreview.net/forum?id=1PL1NIMMrw} {Self-consistency improves chain of thought reasoning in language models}.
\newblock In \emph{International Conference on Learning Representations}.

\bibitem[{Wang et~al.(2024{\natexlab{c}})Wang, Chen, Han, Lin, Zhao, Liu, Zhai, Yuan, You, and Yang}]{wang2024exploringreasoningabilitiesmultimodal}
Yiqi Wang, Wentao Chen, Xiaotian Han, Xudong Lin, Haiteng Zhao, Yongfei Liu, Bohan Zhai, Jianbo Yuan, Quanzeng You, and Hongxia Yang. 2024{\natexlab{c}}.
\newblock \href {https://arxiv.org/abs/2401.06805} {Exploring the reasoning abilities of multimodal large language models (mllms): A comprehensive survey on emerging trends in multimodal reasoning}.
\newblock \emph{Preprint}, arXiv:2401.06805.

\bibitem[{Wei et~al.(2022)Wei, Wang, Schuurmans, Bosma, Ichter, Xia, Chi, Le, and Zhou}]{wei2022chain}
Jason Wei, Xuezhi Wang, Dale Schuurmans, Maarten Bosma, Brian Ichter, Fei Xia, Ed~Chi, Quoc Le, and Denny Zhou. 2022.
\newblock \href {https://proceedings.neurips.cc/paper_files/paper/2022/file/9d5609613524ecf4f15af0f7b31abca4-Paper-Conference.pdf} {Chain-of-thought prompting elicits reasoning in large language models}.
\newblock In \emph{Advances in Neural Information Processing Systems}, volume~35, pages 24824--24837.

\bibitem[{Xu et~al.(2024)Xu, Jin, Li, Song, Sun, and Yuan}]{xu2024llava}
Guowei Xu, Peng Jin, Hao Li, Yibing Song, Lichao Sun, and Li~Yuan. 2024.
\newblock \href {https://arxiv.org/abs/2411.10440} {Llava-cot: Let vision language models reason step-by-step}.
\newblock \emph{arXiv preprint arXiv:2411.10440}.

\bibitem[{Yang et~al.(2024)Yang, Yang, Zhang, Hui, Zheng, Yu, Li, Liu, Huang, Wei, Lin, Yang, Tu, Zhang, Yang, Yang, Zhou, Lin, Dang, Lu, Bao, Yang, Yu, Li, Xue, Zhang, Zhu, Men, Lin, Li, Xia, Ren, Ren, Fan, Su, Zhang, Wan, Liu, Cui, Zhang, and Qiu}]{qwen2.5}
An~Yang, Baosong Yang, Beichen Zhang, Binyuan Hui, Bo~Zheng, Bowen Yu, Chengyuan Li, Dayiheng Liu, Fei Huang, Haoran Wei, Huan Lin, Jian Yang, Jianhong Tu, Jianwei Zhang, Jianxin Yang, Jiaxi Yang, Jingren Zhou, Junyang Lin, Kai Dang, Keming Lu, Keqin Bao, Kexin Yang, Le~Yu, Mei Li, Mingfeng Xue, Pei Zhang, Qin Zhu, Rui Men, Runji Lin, Tianhao Li, Tingyu Xia, Xingzhang Ren, Xuancheng Ren, Yang Fan, Yang Su, Yichang Zhang, Yu~Wan, Yuqiong Liu, Zeyu Cui, Zhenru Zhang, and Zihan Qiu. 2024.
\newblock Qwen2.5 technical report.
\newblock \emph{arXiv preprint arXiv:2412.15115}.

\bibitem[{Yao et~al.(2024)Yao, Huang, Wu, Zhang, Wang, Liu, Wang, Song, Feng, Shen et~al.}]{yao2024mulberry}
Huanjin Yao, Jiaxing Huang, Wenhao Wu, Jingyi Zhang, Yibo Wang, Shunyu Liu, Yingjie Wang, Yuxin Song, Haocheng Feng, Li~Shen, et~al. 2024.
\newblock Mulberry: Empowering mllm with o1-like reasoning and reflection via collective monte carlo tree search.
\newblock \emph{arXiv preprint arXiv:2412.18319}.

\bibitem[{Yao et~al.(2023)Yao, Yu, Zhao, Shafran, Griffiths, Cao, and Narasimhan}]{yao2023tree}
Shunyu Yao, Dian Yu, Jeffrey Zhao, Izhak Shafran, Thomas~L Griffiths, Yuan Cao, and Karthik Narasimhan. 2023.
\newblock \href {https://arxiv.org/abs/2305.10601} {Tree of thoughts: Deliberate problem solving with large language models}.
\newblock \emph{arXiv preprint arXiv:2305.10601}.

\bibitem[{Yue et~al.(2024)Yue, Ni, Zhang, Zheng, Liu, Zhang, Stevens, Jiang, Ren, Sun et~al.}]{yue2024mmmu}
Xiang Yue, Yuansheng Ni, Kai Zhang, Tianyu Zheng, Ruoqi Liu, Ge~Zhang, Samuel Stevens, Dongfu Jiang, Weiming Ren, Yuxuan Sun, et~al. 2024.
\newblock Mmmu: A massive multi-discipline multimodal understanding and reasoning benchmark for expert agi.
\newblock In \emph{Proceedings of the IEEE/CVF Conference on Computer Vision and Pattern Recognition}, pages 9556--9567.

\bibitem[{Zhang et~al.(2023)Zhang, Li, Zhang, Zhang, and Li}]{zhang2023multimodal}
Hanbo Zhang, Xin Li, Yiduo Zhang, Xiaojun Zhang, and Lei Li. 2023.
\newblock \href {https://arxiv.org/abs/2302.00923} {Multimodal chain-of-thought reasoning in language models}.
\newblock \emph{arXiv preprint arXiv:2302.00923}.

\bibitem[{Zhang et~al.(2024{\natexlab{a}})Zhang, Li, Wang, Chen, Zhang, and Chen}]{zhang2024multi}
Xiangyu Zhang, Jiajun Li, Zihan Wang, Xi~Chen, Wen Zhang, and Huajun Chen. 2024{\natexlab{a}}.
\newblock \href {https://www.techrxiv.org/doi/full/10.36227/techrxiv.171259566.60211714/v1} {Multi-modal knowledge graph completion in the wild}.
\newblock \emph{arXiv preprint arXiv:2405.12345}.

\bibitem[{Zhang et~al.(2024{\natexlab{b}})Zhang, Zhang, Zhang, Wang, Li, Zhang, Wang, Zhang, Li, Shi et~al.}]{zhang2024mavis}
Yizhuo Zhang, Yifan Zhang, Yuchen Zhang, Yixin Wang, Yining Li, Zihan Zhang, Yizhuo Wang, Yixin Zhang, Yucheng Li, Yichong Shi, et~al. 2024{\natexlab{b}}.
\newblock \href {https://arxiv.org/abs/2407.08739} {{MAVIS: Mathematical Visual Instruction Tuning}}.
\newblock \emph{arXiv preprint arXiv:2407.08739}.

\bibitem[{Zhao et~al.(2024)Zhao, Zong, Zhang, and Hospedales}]{zhao2024benchmarking}
Bingchen Zhao, Yongshuo Zong, Letian Zhang, and Timothy Hospedales. 2024.
\newblock \href {https://arxiv.org/abs/2406.12742} {Benchmarking multi-image understanding in vision and language models: Perception, knowledge, reasoning, and multi-hop reasoning}.
\newblock \emph{arXiv preprint arXiv:2406.12742}.

\end{thebibliography}

\clearpage
\appendix

\section{Dataset Details}
\label{appendix:dataset}

\begin{figure}[t]
    \centering
    \includegraphics[width=\columnwidth]{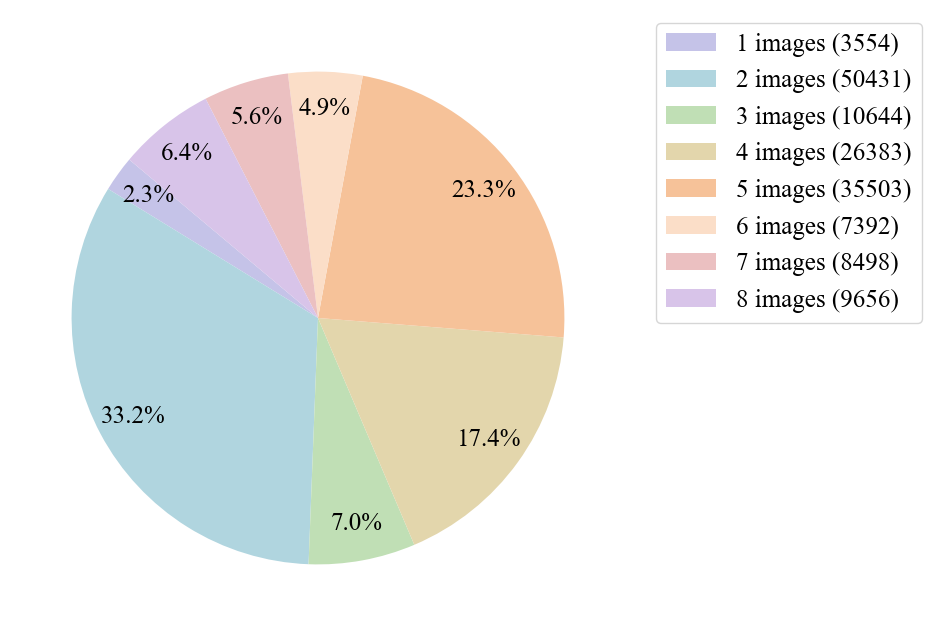}
    \caption{The image count distribution in VISC-150K spans 1–8 images per instance, with samples containing 2–5 images accounting for 80.9\% of the total dataset.}
    \label{image_count}
\end{figure}

VISC-150K comprises 152,061 entries, with images aggregated from publicly accessible websites~\footnote{WikiHow (https://www.wikihow.com), eHow (https://www.ehow.com)} established open-source visual datasets. It includes reasoning paths of lengths 1–8, formatted as open-ended QA and single-choice tasks, and covers varying numbers of input images. The detailed composition is illustrated in Figure \ref{image_count}.

\section{Experimental Settings}
\label{appendix:experimental_settings}
All experiments were conducted on 8 NVIDIA A100 (80GB) GPUs. For training, we applied LoRA fine-tuning with a rank of 16, an initial learning rate of 1e-5, a warmup ratio of 0.05, and a cosine learning rate scheduler. The batch size was set to 8 with a maximum context length of 32,768, using the bfloat16 floating-point format. During inference, we configured the temperature to 0 for deterministic generation, set max new tokens to 1,024, and accelerated computations with FlashAttention-2. Notably, for Qwen2-VL, we dynamically resized input image resolutions to the range of 128×28×28 – 1280×28×28 to balance inference speed and computational resource consumption.

\section{Baseline Details}
\label{appendix:baselines}
\textbf{Qwen2-VL}~\cite{Qwen2VL} incorporates M-RoPE to handle positional information and is capable of processing images of any resolution dynamically. These techniques allow the model to capture detailed visual information more effectively.

\noindent \textbf{Qwen-VL-Chat}~\cite{bai2023qwen} is built upon the foundation of Qwen and is enhanced with visual capabilities through training on high-resolution images and fine-grained datasets.

\noindent \textbf{LLaVA-OneVision}~\cite{li2024llavaonevision} is built upon LLaVA-NeXT series, leveraging large language models (LLMs) and vision encoders connected through a simple projection module. 

\noindent \textbf{LLaVA-1.6}~\cite{liu2024llavanext} addresses the limitations of existing VLMs that primarily focus on single-image tasks. The model leverages an interleaved data format as a general template to unify different visual scenarios. 

\noindent \textbf{LLaVA-1.5}~\cite{liu2023improved} is designed to improve visual reasoning and instruction-following capabilities by leveraging visual instruction tuning. 

\noindent \textbf{InternVL2.5}~\cite{chen2024expanding} is a state-of-the-art VLM built upon the architecture of InternVL2. Its key features include a robust vision encoder, flexible language model integration, dynamic high-resolution processing, and effective test-time scaling.

\noindent \textbf{InternVL2}~\cite{chen2024far} employs a progressive alignment training strategy, allowing the model to scale from smaller to larger sizes while refining the training data from coarse to fine. 

\noindent \textbf{Mantis-Idefics2}~\cite{jiang2024mantis} builds on existing LMM architectures and modifies them to support interleaved multi-image inputs. It uses a text-image interleaving format to mark boundaries between images, enabling the model to better understand and reason over multiple images.

\noindent\textbf{Idefics2}~\cite{laurencon2024idefics2} incorporates learned pooling strategies to reduce the number of visual tokens, significantly improving computational efficiency while maintaining or even enhancing performance.

\noindent \textbf{VILA-1.5}~\cite{lin2023vila} is designed to effectively integrate visual inputs with the strengths of large language models. Interleaved training data and joint SFT enable the model to achieve superior performance on vision-language tasks while retaining strong text-only capabilities.

\noindent \textbf{OpenFlamingo-v2}~\cite{awadalla2023openflamingo} utilizes frozen language models augmented with layers that cross-attend to outputs from a frozen vision encoder. The training on web-scraped image-text sequences enables it to process interleaved sequences of images and text.

\section{Benchmark Details}
\label{appendix:benchmarks}
\noindent \textbf{MMIU}~\cite{meng2024mmiu} categorizes multi-image relationships into three primary types: semantic, spatial, and temporal, grounded in cognitive psychology theory. These categories are further subdivided into seven subtypes, covering 52 distinct multi-image understanding tasks. The benchmark comprises 77K images and 11K multiple-choice questions.

\noindent\textbf{MuirBench}~\cite{wang2024muirbench} consists of 11,264 images and 2,600 multiple-choice questions, covering 12 distinct multi-image understanding tasks and encompassing 10 types of multi-image relationships. By incorporating diverse tasks and image relationships, it establishes a novel and comprehensive benchmark for multi-image understanding.

\noindent \textbf{MIRB}~\cite{zhao2024benchmarking}
includes four evaluation dimensions: 
perceptual understanding, visual world knowledge integration, complex reasoning, and multi-hop reasoning Each dimension incorporates specialized tasks requiring sophisticated cross-image comparison and analytical reasoning.

\noindent \textbf{BLINK}~\cite{fu2024blink} reinterprets traditional computer vision problems as multiple-choice questions, incorporating 14 visual perception tasks that humans can quickly solve. It consists of 3,807 multiple-choice questions accompanied by 7,358 images sourced from multiple datasets, covering indoor, outdoor, and natural scenes.

\noindent \textbf{NLVR2}~\cite{suhr2019nlvr2} focuses on natural language grounding in visual contexts, presenting paired images with corresponding English descriptions. This benchmark emphasizes the evaluation of fine-grained linguistic-visual alignment through diverse reasoning tasks, requiring precise understanding of both textual and visual modalities.

\noindent \textbf{Mantis-Eval}~\cite{jiang2024mantis} comprises 217 high-quality multi-image reasoning samples, covering a range of multi-image skills such as co-reference, reasoning, and comparison.

\noindent \textbf{MVBench}~\cite{li2024mvbench} consists of 20 challenging video understanding tasks that assess temporal reasoning capabilities.

\section{Prompt}
For the four stages of data synthesis, we meticulously crafted structured prompts to achieve the finest-grained task decomposition at each phase. The prompts for each component are demonstrated below:

\begin{tcolorbox}[colback=gray!10,  left=1mm, right=1mm, top=1mm, bottom=1mm] 
\textbf{Feature Extraction:}

You are a visual description expert. Please provide a detailed, comprehensive, and natural language description of the following image, covering every visible detail.

        Overall View:
        - Summarize the scene in 1-2 sentences, focusing on the general setting, lighting, time of day, and the environment. Ensure to include the general mood and ambiance.

        Main Objects:
        - For each key object, describe these aspects in fluent natural language:
            - What is it (e.g., a person, a car, a building)?
            - Quantity, color, size, shape, material, texture, and any distinctive features.
            - Where is it located (foreground, center, background)?
            - State/Function: Is it active or stationary? What is its function in the scene?

        Secondary Objects and Background:
        - Describe smaller or less prominent objects and elements in the background. How do they relate to the main objects? Mention any supporting objects that add depth to the scene (e.g., objects on a table, items in the background, etc.).

        Object Interactions:
        - Highlight interactions or relationships between objects (e.g., people talking, animals interacting). Describe any dynamic actions or static arrangements.

        Text:
        - If there are text in the image, extract all the text and analyze its meanings.

        Atmosphere\&Theme:
        - Convey the mood or theme of the scene, using descriptive adjectives (e.g., lively, serene, chaotic). If unsure, use "seems to" to indicate speculation about the tone.

        Detailed Natural Language Description:
        - Integrate all of the above details into a flowing, cohesive narrative. Ensure to describe every element in fine detail, maintaining clarity and logical structure. Avoid redundancy or skipping any visible detail.

\end{tcolorbox}


\begin{tcolorbox}[colback=gray!10,  left=1mm, right=1mm, top=1mm, bottom=1mm] 
\textbf{Pair Connection:}

You are a professional visual content analyst skilled in analyzing image pairs that exhibit clear correlations.

    User will provide a set of structured descriptions corresponding to images. Based on these descriptions, you are required to analyze the images through an object-oriented or event-oriented approach to identify which image pairs are most strongly correlated. Specifically, you should focus on determining whether there are common objects or associated events/themes between the images. By evaluating the co-occurrence of objects or the relationships between events or topics, return the correlated image pairs as a tuple.
\end{tcolorbox}

\begin{tcolorbox}[colback=gray!10,  left=1mm, right=1mm, top=1mm, bottom=1mm] 
\textbf{Relevance Annotation:}

You are a professional visual content analyst skilled in analyzing relationships between image pairs, including temporal, spatial, and semantic connections.

    User will provide you with two images. Please generate relationship annotations between them based on the following requirements:

    Task Requirements:

    1. Temporal Relationship:
       Identify if there is a clear sequence of events between Image A and Image B. 
         - First, analyze whether the scenes or events in the two images represent a clear chronological order.
         - If there is a clear temporal sequence, describe the progression or transition between the two images, noting the overall process.

    2. Spatial Relationship:
       Analyze if there are any spatial connections or changes in scene or object positions between Image A and Image B.
         - Check if both images depict the same scene or objects in similar layouts.
         - If shared objects or settings are present, compare their positions, orientations, or size differences in both images.

    3. Semantic Relationship: 
       Evaluate if there is a thematic, emotional, or causal connection between Image A and Image B.
         - Determine if the themes, emotional tones, or meaning presented in both images are consistent or related.
         - Assess if there is a cause-and-effect relationship or logical connection between the two images.

    Output Format should be in JSON.
\end{tcolorbox}

\begin{tcolorbox}[colback=gray!10,  left=1mm, right=1mm, top=1mm, bottom=1mm] 
\textbf{Quesiton Generation:}

Task Requirements:
    1. Generate Three Complex Reasoning Questions:
       - Each question should be a multi-step reasoning question, and involve at least three images.
       - Questions should be object-oriented, or event-oriented.
       - Avoid begin with 'How' if possible, and make sure the answer is not open-ended.
       - Questions should be about fine-grained features instead of coarse understanding.
       - Questions types:
         - Detail analysis and comparison
         - Fact judgment
         - Sequence ordering
         - Scene understanding
         - Visual grounding
         - Counterfactual reasoning
         - Action prediction
         - Visual navigation
       - Don't specify images explicitly.
       - Each question must be a single sentence without clauses connected by 'and'.

    2. Decompose Each Complex Question into Sub-Questions and Build a Reasoning Chain:
       - Each sub-question specifies one or two images.
       - Don't focus on the same image twice.
       - Construct a logical reasoning chain for each question, showing the step-by-step connection of sub-questions and answers.

    3. Step-by-Step Answer Each Reasoning Chain to Arrive at the Final Answer

    4. Ensure Data Quality:
       - The questions and answers must be clear, specific, and logically consistent.
       - Avoid irrelevant details or ambiguity, ensuring that all generated content is directly related to the provided image information.

    Output Format should be JSON.
\end{tcolorbox}

\section{Details of Human Annotations}
\begin{table}[t]
\centering
\begin{tabular}{@{}lc@{}}
\toprule
\textbf{Distribution}              & \textbf{Samples} \\ \midrule
Unanimously Correct                & 191               \\
2 Correct vs. 1 Incorrect          & 4                 \\
1 Correct vs. 2 Incorrect          & 2                 \\
Unanimously Incorrect              & 3                 \\ \bottomrule
\end{tabular}
\caption{Statistical analysis of human annotation results conducted on 200 sampled instances. Each instance's annotation outcome was classified into four categories: (1) unanimously correct, (2) two annotators rated as correct vs. one incorrect, (3) two annotators rated as incorrect vs. one correct, and (4) unanimously incorrect. A data instance is considered valid if at least two annotators labeled it as correct; otherwise, it is deemed invalid.}
\label{table: annatation}
\end{table}

We recruited three annotators (Ph.D. students in Computer Science and Technology) to evaluate the correctness of 200 randomly sampled instances from VISC-150K. For each instance, the evaluation criteria included three dimensions:

\noindent$\bullet$ Final Answer Correctness: Whether the annotator deemed the final answer to accurately resolve the original question;

\noindent$\bullet$ Sub-Question Answer Correctness: Whether intermediate answers correctly addressed their corresponding sub-questions;

\noindent$\bullet$ Visual Focus Validity: Whether the visual focus at each step aligned with the minimal essential visual information required for reasoning.
\begin{figure*}[ht]
    \centering
    \includegraphics[width=\textwidth]{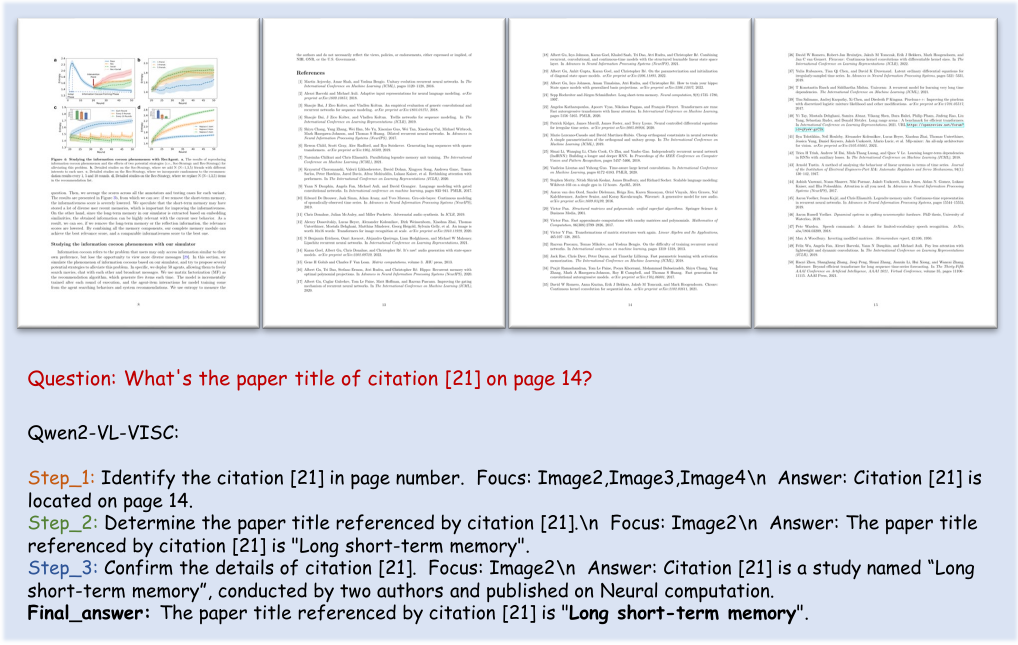}
    \caption{In the cross-image visual reasoning case, Qwen2-VL-VISC trained on the VISC-150K dataset with the acquired \textit{Focus-Centric Visual Chain} reasoning paradigm accurately resolves the reasoning task.}
    \label{case_study_1}
\end{figure*}

\begin{figure*}[ht]
    \centering
    \includegraphics[width=\textwidth]{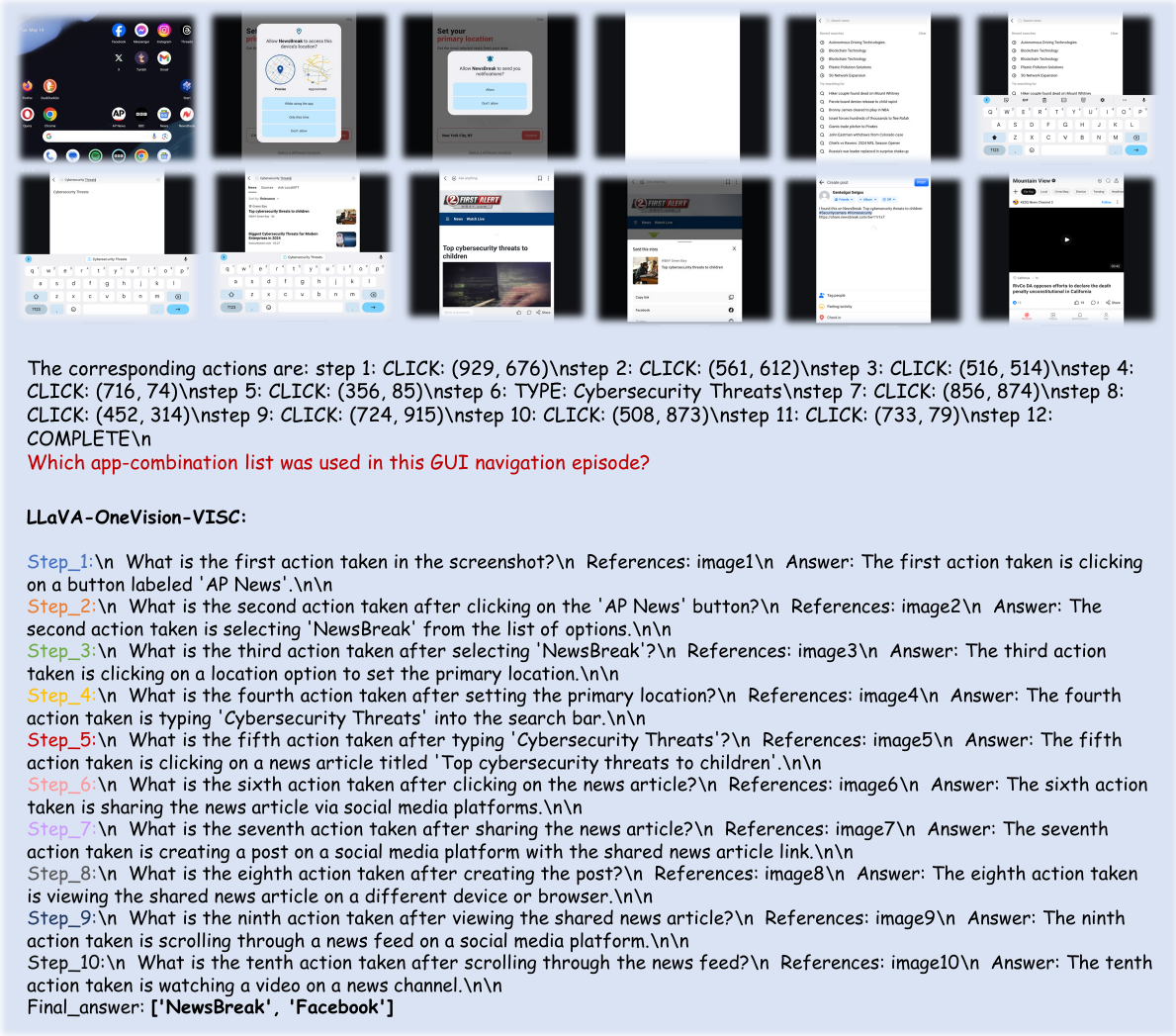}
    \caption{In a cross-image GUI navigation case study, LLaVA-OneVision-VISC, guided by the \textit{Focus-Centric Visual Chain} paradigm, constructs a multi-step reasoning trajectory. By iteratively focusing on GUI elements within individual images at each stage, it deduces a comprehensive operational workflow and ultimately arrives at the correct solution.}
    \label{case_study_2}
\end{figure*}

An instance was labeled as correct by an annotator only when all three criteria were satisfied. The annotation results from the three annotators are summarized in Table \ref{table: annatation}. A data instance was marked as valid if at least two annotators labeled it as correct; otherwise, it was deemed invalid. Based on the evaluation results, our synthetic data achieved a 97.5\% validity rate under substantial inter-annotator agreement (Fleiss' $\kappa$ = 0.637).
s
\section{Case Study}
We present test cases for Qwen2-VL-VISC and LLaVA-OneVision-VISC fine-tuned on VISC-150K, illustrated in Figure~\ref{case_study_1} and Figure~\ref{case_study_2} respectively. These cases effectively demonstrate the multi-image reasoning and integrated information processing capabilities of our data-augmented models.

\end{document}